\documentclass[journal]{IEEEtran}
\usepackage{verbatim}

\usepackage{cite}

\ifCLASSINFOpdf
   \usepackage[pdftex]{graphicx}
   \graphicspath{{./images/}}
\else
\fi
\usepackage[cmex10]{amsmath}
\usepackage{amssymb}
\usepackage{latexsym}
\usepackage{psfrag}
\usepackage{epsfig}
\usepackage{graphicx}
\usepackage{epstopdf}
\usepackage{multicol}
\usepackage{arydshln}
\usepackage{color}
\usepackage{tikz}
\usetikzlibrary{calc}

\newtheorem{theorem}{\textbf{Theorem}}[section]
\newtheorem{lemma}[theorem]{\textbf{Lemma}}
\newtheorem{proposition}[theorem]{\textbf{Proposition}}

\newtheorem{example}[theorem]{\textbf{Example}}

\newtheorem{remark}[theorem]{Remark}
\newtheorem{algorithm}[theorem]{\textbf{Algorithm}}
\newtheorem{assumption}[theorem]{\textbf{Assumption}}

\newcommand{\R}{{\rm I\!R}}
\newcommand{\dfb}{\stackrel{\Delta}{=}}
\newcommand{\sbm}[1]{\left[\begin{smallmatrix} #1
   \end{smallmatrix}\right]}

\usepackage{caption}
\usepackage{subcaption}
\usepackage{url}

\begin{document}
%
\title{Distributed rotational and translational maneuvering of rigid formations and their applications.}
%
%
%
\author{Hector~Garcia de Marina,~\IEEEmembership{Member,~IEEE,}
	Bayu~Jayawardhana,~\IEEEmembership{Senior Member,~IEEE,}
        and Ming~Cao,~\IEEEmembership{Senior Member,~IEEE}%
\thanks{The authors are with the Engineering and Technology Institute of Groningen, University of Groningen, 9747 AG Groningen, The Netherlands. (e-mail: \{h.j.de.marina, b.jayawardhana, m.cao\}@rug.nl). This work was supported by the the EU INTERREG program under the auspices of the SMARTBOT project and the work of Cao was also supported by the European Research Council
(ERC-StG-307207).}}


%
%

\markboth{IEEE TRANSACTIONS ON ROBOTICS}%
{}
%



\maketitle

\begin{abstract}
	Recently it has been reported that range-measurement inconsistency, or equivalently mismatches in prescribed inter-agent distances, may prevent the popular gradient controllers from guiding rigid formations of mobile agents to converge to their desired shape, and even worse from standing still at any location. In this paper, instead of treating mismatches as the source of ill performance, we take them as design parameters and show that by introducing such a pair of parameters per distance constraint, distributed controller achieving simultaneously both formation and motion control can be designed that not only encompasses the popular gradient control, but more importantly allows us to  achieve constant collective translation, rotation or their combination while guaranteeing asymptotically no distortion in the formation shape occurs. Such motion control results are then applied to (a) the alignment of formations' orientations and (b) enclosing and tracking a moving target. Besides rigorous mathematical proof, experiments using mobile robots are demonstrated to show the satisfying performances of the proposed formation-motion distributed controller.
\end{abstract}

\begin{IEEEkeywords}
	Formation Control, Rigid Formation, Motion Control, Target enclosing.
\end{IEEEkeywords}

%

\section{Introduction}
%
%
%
%
\IEEEPARstart{M}{any} coordinated robot tasks, such as, the enclosing of a target \cite{LanYanLin10}, area exploration \& surveillance \cite{Yuan10} and the vehicle platooning for energy efficiency \cite{MaStGrInSp08}, can be achieved by combining two different cooperative controls: multi-agent formation control and group motion control. The former is  to achieve and maintain a specific desired shape of a multi-agent formation while the latter to guide the motion of the group as a whole. For simple formations, such as, line formations in the vehicle platooning problem, these two control problems can be solved simultaneously by using the passivity-based control approach as pursued in \cite{Vos12}. However, for formations in more complicated shapes, these two control problems are usually tackled separately, namely using the gradient-based strategies for formation control and leader-follower coordination for motion control, where for the latter the leader moves according to a desired trajectory and the followers simply track the leader \cite{BaArWe11}. But when the separately designed formation controller and motion controller are jointly in force as in the common scenario in practice, conflict often occurs: the shape of the formation is distorted when an agent compromises between the demands from formation and motion controls as to where and how it needs to move \cite{Skj02,antonelli2009experiments}. While efforts have been made before to solve such conflict by installing velocity sensors in addition to position sensors on  mobile agents and thus compensating the distortion with the help of the more sensed information \cite{BaArWe11}, we in this paper take a much more head-on approach by generalizing the gradient-based distributed formation control law for both control objectives simultaneously guaranteeing no shape distortion in the steady-state desired formation motion even without the knowledge of any additional sensed information in comparison to the conventional control. The surprisingly simple structure of our proposed control law opens possibilities to solve other difficult problems, such as, collective rotational motion, the enclosing of a moving target and the formation coordination task for agents governed by higher-order dynamics.

In gradient-based formation control, stabilizable formations are identified employing rigidity graph theory where the vertices of a graph represent the agents and the edges stand for the inter-agent distance constraints to define the shape of the formation \cite{OlMu02,KrBrFr08,YuAnDaFi09,CaYuAn11}. Rigid formations are associated with a potential function determined by the agents' relative positions. The potential has a minimum at the desired distances between the agents and thus its gradient leads naturally to the formation controller that stabilizes rigid formations locally. However, it was recently reported in \cite{BeMoMoAn12} and later more exhaustively studied in \cite{MouMorseBelSunAnd15,Hem14} that for such control laws, if there are mismatches between the prescribed distances for neighboring agents, the performance of the controller deteriorates significantly since mismatches induce undesired steady-state motion and distortion in the formation shape. If no estimators, as proposed and experimented in \cite{MarCaoJa15}, are installed to improve the performance of such controllers, the mismatch-induced motion in steady-states takes interesting forms for almost all mismatches: in $\R^2$, the formation follows a closed orbit with a constant angular velocity; in $\R^3$ the movement is helical, determined by a single sinusoidal signal and a constant drift \cite{SunMouAndMor14}.

Motivated by the mismatch-induced motion reported in \cite{SunMouAndMor14}, we introduce parameter pairs in the gradient-based formation control law for steering the whole formation; in particular, for each prescribed distance constraint, we introduce a pair of motion parameters, mimicking the mismatches as studied in \cite{SunMouAndMor14}. By a one-step (centralized) systematic design of such motion parameters, we show that we can steer the entire formation in rotation and/or translation in a distributed manner. The proposed gradient-based formation-motion joint controller using the motion parameters has several advantages. First, one can achieve precisely the desired steady-state formation shape and the group motion simultaneously, including rotational motion, in a distributed fashion. To the best of our knowledge, no such result has been reported in the literature. For single-integrator agent dynamics, we relate analytically the magnitude of the motion parameters to the speed of the group motion. Second, our proposed approach enables us to align the formation with respect to a global coordinate frame by adding a simple control term to an arbitrary agent when guiding the group in motion. In comparison, in the literature such an alignment is usually obtained by assigning a leader who knows the global frame and letting the other agents (followers) estimate the leader's velocity employing a distributed estimator \cite{BaArWe11}. Obviously in our approach, such estimators are not needed at all. Third, one can easily use our method to address the target tracking and enclosing problem both in $\R^2$ and $\R^3$ where one assigns the formation shape to specify how the target is enclosed and tracked by the pursuers. In comparison to the solution of such problems in \cite{Guo10,Mar12} among other works, we do not confine our formation to follow the same circular trajectory nor require \emph{all} enclosing agents to measure their relative positions with respect to the target.
The rest of the paper is organized as follows. In Section \ref{sec: pre} we introduce the notation and background for rigid formations in $\R^2$ and $\R^3$ and we briefly review the gradient control in Section \ref{sec: grad}. The design of the desired steady state motion with the desired formation shape  is described in Section \ref{sec: motion}, while the corresponding stability analysis  is presented in Section \ref{sec: sta}. We implement our proposed formation-motion controller in two different contexts: (i) the design of a simultaneous formation \& translation control in global coordinates in Section \ref{sec: tra}; and (ii) the formation control design for solving the target tracking and enclosing problem in Section \ref{sec: enc}. Finally experimental results using wheeled mobile robots are shown in Section \ref{sec: exp}.

\section{Preliminaries}
\label{sec: pre}
In this section, we introduce some notations and basic concepts. For a given matrix $A\in\R^{n\times p}$, define $\overline A \dfb A \otimes I_m \in\R^{nm\times pm}$, where the symbol $\otimes$ denotes the Kronecker product, $m = 2$ for $\R^2$ or otherwise $3$ for $\R^3$, and $I_m$ is the $m$-dimensional identity matrix. For a stacked vector $x\dfb \begin{bmatrix}x_1^T & x_2^T & \dots & x_k^T\end{bmatrix}^T$ with $x_i\in\R^{n}, i\in\{1,\dots,k\}$, we define the diagonal matrix $D_x \dfb \operatorname{diag}\{x_i\}_{i\in\{1,\dots,k\}} \in\R^{kn\times k}$. We denote by $|\mathcal{X}|$ the cardinality of the set $\mathcal{X}$ and by $||x||$ the Euclidean norm of a vector $x$. We use $\mathbf{1}_{n\times m}$ and $\mathbf{0}_{n\times m}$ to denote the all-one and all-zero matrix in $\R^{n\times m}$ respectively.

\subsection{Formations and graphs}
\label{sec: preA}
We consider a formation of $n\geq 2$ autonomous agents whose positions are denoted by $p_i\in\R^m$. The agents are able to sense the relative positions of its neighboring agents. The neighbor relationships are described by an undirected graph $\mathbb{G} = (\mathcal{V}, \mathcal{E})$ with the vertex set $\mathcal{V} = \{1, \dots, n\}$ and the ordered edge set $\mathcal{E}\subseteq\mathcal{V}\times\mathcal{V}$. The set $\mathcal{N}_i$ of the neighbors of agent $i$ is defined by $\mathcal{N}_i\dfb\{j\in\mathcal{V}:(i,j)\in\mathcal{E}\}$. We define the elements of the incidence matrix $B\in\R^{|\mathcal{V}|\times|\mathcal{E}|}$ for  $\mathbb{G}$ by
\begin{equation*}
	b_{ik} \dfb \begin{cases}+1 \quad \text{if} \quad i = {\mathcal{E}_k^{\text{tail}}} \\
		-1 \quad \text{if} \quad i = {\mathcal{E}_k^{\text{head}}} \\
		0 \quad \text{otherwise}
	\end{cases},
\end{equation*}
where $\mathcal{E}_k^{\text{tail}}$ and $\mathcal{E}_k^{\text{head}}$ denote the tail and head nodes, respectively, of the edge $\mathcal{E}_k$, i.e. $\mathcal{E}_k = (\mathcal{E}_k^{\text{tail}},\mathcal{E}_k^{\text{head}})$. A \emph{framework} is defined by the pair $(\mathbb{G}, p)$, where $p = \operatorname{col}\{p_1, \dots, p_n\}$ is the stacked vector of the agents' positions $p_i,i\in\{1,\dots,n\}$. The stacked vector of the sensed relative distances can then be described by
\begin{equation*}
	z = \overline B^Tp.
\end{equation*}
Note that each vector $z_k = p_i - p_j$ in $z$ corresponds to the relative position associated with the edge $\mathcal{E}_k = (i, j)$.

\begin{figure}
	\centering
	\begin{subfigure}{0.2\columnwidth}
\begin{tikzpicture}[line join=round]
\draw(0,0)--(1,0)--(1,1)--(0,1)--(0,0);
\filldraw(0,0) circle (2pt);
\filldraw(1,0) circle (2pt);
\filldraw(1,1) circle (2pt);
\filldraw(0,1) circle (2pt);
\end{tikzpicture}
		\caption{}
	\end{subfigure}
	\begin{subfigure}{0.2\columnwidth}
\begin{tikzpicture}[line join=round]
\filldraw(0,0) circle (2pt);
\filldraw(0,1) circle (2pt);
\filldraw(.5,.5) circle (2pt);
\filldraw(1,.5) circle (2pt);
\filldraw(1.5,1) circle (2pt);
\filldraw(1.5,0) circle (2pt);
\draw(0,0)--(0,1);
\draw(0,1)--(.5,.5);
\draw(.5,.5)--(0,0);
\draw(1,.5)--(1.5,1);
\draw(1.5,1)--(1.5,0);
\draw(1.5,0)--(1,.5);
\draw(0,1)--(1.5,1);
\draw(.5,.5)--(1,.5);
\draw(0,0)--(1.5,0);
\end{tikzpicture}
		\caption{}
	\end{subfigure}
	\begin{subfigure}{0.2\columnwidth}
\begin{tikzpicture}[line join=round]
\filldraw(0,0) circle (2pt);
\filldraw(.5,.5) circle (2pt);
\filldraw(1,1) circle (2pt);
\draw(0,0)--(.5,.5);
\draw(.5,.5)--(1,1);
\draw(-.02,.06)--(.98,1.06);
\end{tikzpicture}
		\caption{}
	\end{subfigure}
	\begin{subfigure}{0.2\columnwidth}
\begin{tikzpicture}[line join=round]
\filldraw[fill=white](0,0)--(1,1)--(0,1)--cycle;
\filldraw(0,0) circle (2pt);
\filldraw(1,1) circle (2pt);
\filldraw(0,1) circle (2pt);
\end{tikzpicture}
		\caption{}
	\end{subfigure}
	\begin{subfigure}{0.2\columnwidth}
\begin{tikzpicture}[line join=round]
[\tikzset{>=latex}]\filldraw[draw=black,fill=white,fill opacity=0.8](-.132,-.005)--(.275,-.23)--(-.275,-.396)--(-.683,-.171)--cycle;
\filldraw[draw=black,fill=white,fill opacity=0.8](-.132,.62)--(-.132,-.005)--(-.683,-.171)--(-.683,.454)--cycle;
\filldraw[draw=black,fill=white,fill opacity=0.8](-.132,-.005)--(-.132,.62)--(.275,.396)--(.275,-.23)--cycle;
\filldraw[draw=black,fill=white,fill opacity=0.8](.275,.396)--(-.132,.62)--(-.683,.454)--(-.275,.23)--cycle;
\filldraw[draw=black,fill=white,fill opacity=0.8](-.275,-.396)--(-.275,.23)--(-.683,.454)--(-.683,-.171)--cycle;
\filldraw[draw=black,fill=white,fill opacity=0.8](.275,-.23)--(.275,.396)--(-.275,.23)--(-.275,-.396)--cycle;
\end{tikzpicture}
		\caption{}
	\end{subfigure}
	\begin{subfigure}{0.2\columnwidth}
\begin{tikzpicture}[line join=round]
[\tikzset{>=latex}]\filldraw[draw=black,fill=white,fill opacity=0.8](-.182,-.171)--(0,0)--(0,0)--(-.272,.15)--cycle;
\filldraw[draw=black,fill=white,fill opacity=0.8](-.272,.15)--(0,0)--(0,0)--(.454,.021)--cycle;
\draw[arrows=<->,thick](-.367,-.111)--(-.639,.039)--(-.272,.15);
\draw[arrows=->,thick](-.639,.039)--(-.639,.456);
\filldraw[draw=black,fill=white,fill opacity=0.8](.454,.021)--(0,0)--(0,0)--(-.182,-.171)--cycle;
\end{tikzpicture}
		\caption{}
	\end{subfigure}
	\begin{subfigure}{0.2\columnwidth}
\begin{tikzpicture}[line join=round]
[\tikzset{>=latex}]\filldraw[draw=black,fill=white,fill opacity=0.8](-.611,.337)--(-.204,.738)--(-.204,.738)--(.477,.144)--cycle;
\filldraw[draw=black,fill=white,fill opacity=0.8](-.477,-.144)--(-.204,.738)--(-.204,.738)--(-.611,.337)--cycle;
\filldraw[draw=black,fill=white,fill opacity=0.8](.477,.144)--(-.204,.738)--(-.204,.738)--(-.477,-.144)--cycle;
\end{tikzpicture}
		\caption{}
	\end{subfigure}
	\caption{a) The square without an inner diagonal is not rigid since we can smoothly move the top two nodes keeping the other two fixed without breaking the distance constraints; b) A rigid but not an infinitesimally rigid framework. If we rotate the left inner triangle, then the right inner triangle can be counter-rotated  in order to keep the inter-distances constant; c) A minimally rigid but not an infinitesimally rigid framework since the nodes' positions are collinear; d) The triangle is infinitesimally and minimally rigid; e) The cube formed by squares without diagonals is not rigid; f) The zero-volume tetrahedron is rigid but not infinitesimally rigid  in $\R^3$, since all the nodes are co-planar; and g) The tetrahedron in $\R^3$ is infinitesimally and minimally rigid.}
	\label{fig: rigid}
\end{figure}
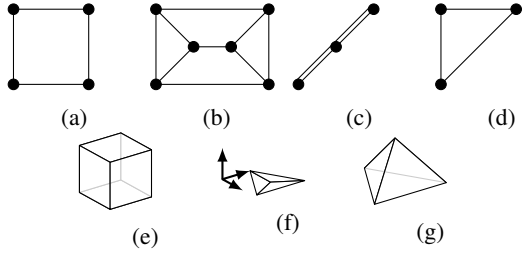

{
\subsection{Infinitesimally and minimally rigid formations and their realization}
In this section we will briefly review the concept of rigid formations and how to design them. Most of the concepts explained here are covered in much more detail in \cite{AsRo79} and \cite{AnYuFiHe08}. Define the edge function $f_\mathbb{G}$ by $f_{\mathbb{G}}(p) = \mathop{\text{col}}\limits_{k}\big(\|z_k\|^2\big)$ and denote its Jacobian by $R(z)=D_z^T\overline B^T$, which is called in the literature the {\it rigidity matrix}. A framework $(\mathbb{G}, p)$ is {\it infinitesimally rigid} if $\text{rank} R(z) = 2n-3$ when embedded in $\mathbb{R}^2$ or if $\text{rank} R(z) = 3n-6$ when it is embedded in $\mathbb{R}^3$. Additionally, if $|\mathcal E|=2n-3$ in the 2D case or $|\mathcal E|=3n-6$ in the 3D case then the framework is called {\it minimally rigid}. Roughly speaking, under the distance contraints the only motions that one can perform over the agents in an infinitesimally and minimally rigid framework, while they are already in the desired shape, are the ones defining translations and rotations of the whole shape. We illustrate in Figure \ref{fig: rigid} some examples in $\R^2$ and $\R^3$ of rigid and non-rigid frameworks.

}

For a given stacked vector of desired relative positions $z^* = [\begin{smallmatrix}{z_1^*}^T & {z_2^*}^T & \dots & {z_{|\mathcal{E}|}^*}^T\end{smallmatrix}]^T$, the resulting set $\mathcal{Z}$ of the possible
formations is defined by
\begin{equation*}
	\mathcal{Z} \dfb \left \{\left(I_{|\mathcal{E}|} \otimes \mathcal{R}\right)z^* \right \}, 
\end{equation*}
where $\mathcal{R}$ is the set of rotational matrices in $\R^2$ or $\R^3$. 
Roughly speaking, $\mathcal{Z}$ consists of all formation positions that are obtained by rotating $z^*$. If $(\mathbb G, p)$ is infinitesimally and minimally rigid, then, similar to the above, we can define the set of resulting formations $\mathcal{D}$ by
\begin{align}
\mathcal{D} \dfb & \Big \{z  \, | \, ||z_k||=d_k , k\in\{1, \dots, |\mathcal{E}|\} \Big \}, \nonumber 
\end{align}
where $d_k = ||z_k^*||, k\in\{1, \dots, |\mathcal{E}|\}$.

Note that in general it holds that $\mathcal{Z}\subseteq\mathcal{D}$. For a desired shape, one can always design $\mathbb G$ to make the formation infinitesimally and minimally rigid. In fact in $\R^2$, an infinitesimally and minimally rigid framework with two or more vertices can always be constructed through the Henneberg construction \cite{Hen11}. Unfortunately there is no known corresponding systematic operations  in $\R^3$ yet. Nevertheless in $\R^3$ one can always obtain an infinitesimally and minimally rigid framework by the Henneberg-like \emph{tetrahedron insertions} as it has been done in \cite{MarCaoJa15}.

\section{Gradient control in inter-agent distances} \label{sec: grad}
For a formation of $n$ agents associated with the neighbor relationship graph $\mathbb G$, consider the following system where we model the agents as kinematical points
\begin{equation}
	\dot p = u,
	\label{eq: p}
\end{equation}
where $u$ is the stacked vector of control inputs $u_i\in\R^m$ for $i=\{1,\dots,n\}$. 

For each edge $\mathcal{E}_k$ one can construct a potential function $V_k$ with its minimum at the desired distance $||z_k^*||$ so that the gradient of such functions can be used to control inter-agent distances distributively. 

We take as a potential function 
\begin{equation*}
	V\left(\overline B^Tp\right) = V(z) = \sum_{k=1}^{|\mathcal{E}|} V_k(z_k),
\end{equation*}
and apply to each agent $i$ in (\ref{eq: p}) the following gradient descent control
\begin{equation}
	u_i = -\nabla_{p_i}\sum_{k=1}^{|\mathcal{E}|} V_k(z_k).
	\label{eq: pdyn}
\end{equation}
One can then show straightforwardly that the multi-agent formation will converge locally to the desired shape employing the following known results.
\vspace{2pt}
\begin{theorem}
	\cite{Ab06,Oh14} If the potential $V(z)$ is real analytic, i.e. it possesses derivatives of all orders and agrees with its Taylor series in the neighborhood of every point, then the formation under the distributed (\ref{eq: pdyn}) will converge locally asymptotically to the desired shape $\mathcal{Z}$.
	\label{th: V}
\end{theorem}
\vspace{2pt}

The authors of \cite{Ab06,Oh14} have shown that for a real analytic potential function $V$,  system (\ref{eq: p}) with the control law (\ref{eq: pdyn}) is locally asymptotically stable about the point corresponding to the local minima of $V(z)$. We consider for each edge $\mathcal{E}_k$ the following family of quadratic\footnote{Any analytic potential functions, including the ones for collision avoidance \cite{TiWa11}, can be approximated by a quadratic function in a small neighborhood about their local minima, and for this reason such quadratic potentials are of special interest of study.} potential functions
\begin{equation}
	V_k(||z_k||) = \frac{1}{2l}(||z_k||^l - d_k^l)^2,\quad l\in\mathbb{N},
	\label{eq: Vkquad}
\end{equation}
with the following gradient along $z_k$
\begin{align}
	\nabla_{z_k}V_k(||z_k||) &= (\nabla_{z_k}||z_k||)\left(\frac{\partial V_k(||z_k||)}{\partial ||z_k||}\right) \nonumber \\
	&= z_k ||z_k||^{l-2}(||z_k||^l - d_k^l).
	\label{eq: zgrad}
\end{align}
One immediate observation is that when $l=1$, the right-hand side of (\ref{eq: zgrad}) becomes the unit vector $\frac{z_k}{||z_k||}$ multiplied by the distance error  $(||z_k|| - d_k)$. This fact will play an important role later for the design of the desired steady-state motion of the formation. It is straightforward to check that for every edge $\mathcal{E}_k=(i,j)$ one has $\nabla_{p_i}V_k = -\nabla_{p_j}V_k = \nabla_{z_k}V_k$, and obviously $\nabla_{p_h}V_k = \mathbf{0}_{m\times 1}$ for all $h\neq i,j$. Thus the agent dynamics (\ref{eq: p}) under (\ref{eq: pdyn}) can be written in the following compact form
\begin{equation}
	\dot p = -\overline B\nabla_zV(z),
	\label{eq: pB}
\end{equation}
where $\nabla_zV(z)$ is the stacked vector of $\nabla_{z_k}V_k(||z_k||)$'s.
Denoting the distance error for edge $k$ by
\begin{equation}
	e_k = ||z_k||^l - d_k^l,
	\label{eq: ek}
\end{equation}
it follows that
\begin{equation*}
	\nabla_{z_k}V_k(||z_k||) = z_k ||z_k||^{l-2}e_k.
\end{equation*}
By substituting it into (\ref{eq: pB}) and  noting that
\begin{align}
	\dot e_k &= l\,||z_k||^{l-1}\frac{d}{dt}||z_k|| = l\,||z_k||^{l-2}z_k^T\dot z_k \nonumber,
\end{align}
we can write down the closed-loop system dynamics in the compact form
\begin{align}
	\dot p &= -\overline BD_zD_{\tilde z}e = -R(z)^TD_{\tilde z}e \label{eq: pdynNOmu}\\
	\dot z &= \overline B^T\dot p  = -\overline B^TR(z)^TD_{\tilde z}e \label{eq: zdynNOmu} \\
	\dot e &= lD_{\tilde z}D_z^T\dot z = -lD_{\tilde z}R(z)R(z)^TD_{\tilde z}e, \label{eq: edyn}
\end{align}
where $e\in\R^{|\mathcal{E}|}$ is the stacked vector of $e_k$'s, $\tilde z\in\R^{|\mathcal{E}|}$ is the stacked column vector consisting of all the $||z_k||^{l-2}$'s, and the matrices $D_z$ and $D_{\tilde z}$ are the block diagonal matrices of $z$ and $\tilde z$ respectively, as defined before in Section \ref{sec: preA}.
In the following proposition, we establish the local exponential convergence of the error  (\ref{eq: ek}) to zero for arbitrary $l\in\mathbb{N}$. This
fact is a straightforward extension of the result in \cite{MouMorseBelSunAnd15} for the case when $l=2$, and we provide the proof below for completeness.
\begin{proposition}
	\label{pro: staE}
	Consider the closed-loop system (\ref{eq: pdynNOmu})-(\ref{eq: edyn}) with graph $\mathbb{G}$ driven by the gradient of the potential function $V = \sum_k V_k$ where $V_k$ is defined in (\ref{eq: Vkquad}) for a fixed $l\in\mathbb{N}$. If the formation in the desired shape $\mathcal{D}$ is infinitesimally and minimally rigid, then the error signal $e$ goes to zero locally exponentially fast.
\end{proposition}
\begin{IEEEproof}
	First we prove that the error system (\ref{eq: edyn}) is autonomous. The elements of $R(z)R(z)^T = D_z^T\overline B^T\overline BD_z$ are products of the form $z_i^Tz_j,\, i,j\in\{1, \dots, |\mathcal{E}|\}$. It has been shown in \cite{Hem14} that for any infinitesimally and minimally rigid framework, there exists a neighborhood $\mathcal{U}_z$ about this framework such that any $z_i^Tz_j, \, z_i,z_j\in\mathcal{U}_z$, can be written as a smooth function $g_{ij}(||z_k||^2)$, and since $||z_k|| = (e_k + d_k^l)^{\frac{1}{l}}$,  we can write
	\begin{equation}
		z^T_iz_j = g_{ij}(e).
		\label{eq: gij}
	\end{equation}
	Thus the vector $\tilde z$ and the matrix $R(z)R(z)^T$ can be rewritten as smooth functions depending locally only on $e$ about the origin. This shows that the error system (\ref{eq: edyn}) is at least locally an autonomous system. 
	
	Now to prove the stability of $e=0$ for (\ref{eq: edyn}), we choose a Lyapunov function $V = \frac{1}{2l}||e||^2$ which satisfies
	\begin{align}
		\frac{\mathrm{d}V}{\mathrm{d}t} &= \frac{1}{l}e^T\dot e	= -e^TD_{\tilde z}R(z)R(z)^TD_{\tilde z}e \nonumber 
	\end{align}
Since  $R(z)R(z)^T$ and $D_{\tilde z}$ are positive definite matrices at the desired formation shape as $R(z)$ has full row rank at $z\in\mathcal{D}$, the matrix 
	\begin{equation}
		Q(e) = D_{\tilde z}R(z)R(z)^TD_{\tilde z}
		\label{eq: Qe}
	\end{equation}
	is positive definite at $e = 0$. Since the eigenvalues of $Q(e)$ are smooth functions of $e$, there exists a compact set 
	\begin{equation}
		\mathcal{Q}\dfb\{e : ||e||^2\leq \rho\}
		\label{eq: Qset}
	\end{equation}
	for some positive constant $\rho$ where $Q(e)$ is positive definite. Define $\lambda_{\text{min}}$ to be the smallest eigenvalue of $Q(e)$ in $\mathcal{Q}$. Then
	\begin{equation}
		\frac{\mathrm{d}V}{\mathrm{d}t} \leq -\lambda_{\text{min}} ||e||^2,
		\label{eq: lam}
	\end{equation}
	which immediately implies the local exponential convergence of $e$ to zero.
\end{IEEEproof}
We want to emphasize that $e(t)\to 0$ as $t\to\infty$ does not imply $z(t)\to\mathcal{Z}$ but instead  $z(t)\to\mathcal{D}$. As pointed out by Theorem \ref{th: V}, $z(t)$ is guaranteed to converge to $\mathcal{Z}$ only when the initial condition $z(0)$ is sufficiently close to $\mathcal{Z}$ .

\section{Gradient-based formation-motion control}
\label{sec: motion}
For the sake of simplicity and clarity in the notation, from this point onwards we will consider the case for $l=2$ and the main results can be easily extended to any $l\in\mathbb{N}$. Then $\tilde z = \mathbf{1}_{|\mathcal{E}|\times 1}$ and thus $D_{\tilde z}$ is the identity matrix. We denote by $O_g$  the \emph{global frame} of coordinates fixed at the origin of $\R^m$ with some arbitrary fixed orientation. Similarly we denote by $O_i$  the \emph{local frame}  for  agent $i$ with some fixed arbitrary orientation independent of $O_g$. Finally, we denote by $O_b$  the \emph{body frame}  fixed at the centroid $p_c$ of the desired rigid formation. Furthermore, if we rotate the rigid formation with respect to $O_g$, then $O_b$ is also rotated in the same manner. Let $^ip_j$ denote the position of agent $j$ with respect to $O_i$. To simplify notation whenever we represent an agents' variable with respect to $O_g$, the superscript is omitted, e.g. $p_j \dfb {^gp_j}$.

We consider a pair of (scaled by a gain $c\in\R$) motion-control parameters $\frac{\mu_k}{c}\in\R$ and $\frac{\tilde\mu_k}{c}\in\R$ to the term $d_k^2$ for each agent associated with the edge $\mathcal{E}_k=(i,j)$. More precisely, agent $i$ uses a \emph{controlled} distance $d_k^2 + \frac{\mu_k}{c}$ and correspondingly, agent $j$, $d_k^2 - \frac{\tilde\mu_k}{c}$. For the corresponding edge $\mathcal{E}_k=(i,j)$ we incorporate these parameters to the gradient descent controller and apply the same gain $c$, namely
\begin{align}
	u_i^k &= -cz_k (||z_k||^2 - d_k^2)	+ \mu_k z_k \nonumber \\
	u_j^k &= cz_k (||z_k||^2 - d_k^2)  + \tilde\mu_k z_k, \nonumber
\end{align}
where $u_i^k$ and $u_j^k$ are the corresponding control inputs for agents $i$ and $j$ with edge $\mathcal{E}_k$. The equations above lead to the following compact form
\begin{align}
	\dot p &= -cR(z)^Te + \overline A(\mu, \tilde\mu)z,
	\label{eq: pdynCMU}
\end{align}
where the elements $a_{ik}$ of $A$ are constructed in a very similar way as in the incidence matrix as follows
\begin{equation}
	a_{ik} \dfb \begin{cases}\mu_k \quad \text{if} \quad i = {\mathcal{E}_k^{\text{tail}}} \\
		\tilde\mu_k \quad \text{if} \quad i = {\mathcal{E}_k^{\text{head}}} \\
						  0 \quad \text{otherwise,}
					\end{cases}
					\label{eq: A}
\end{equation}
therefore we define $\mu, \tilde\mu\in\R^{|\mathcal{E}|}$ as the stacked vectors of $\mu_k$ and $\tilde\mu_k$ for all $k\in\{1,\dots,|\mathcal{E}|\}$. Note that if $\mu = -\tilde\mu$, then $d^2_k + \frac{\mu_k}{c} = d^2_k - \frac{\tilde\mu_k}{c} = \tilde d^2_k$, which implies that we have a gradient-based control with $\tilde d^2_k$ being the new stacked vector of the prescribed distances. The gain $c$ is a free design parameter for achieving exponential stability of the formation as we will see later in the stability analysis.

One important property of (\ref{eq: pdynCMU}) is that each agent $i$ can work with only its own local frame $O_i$. One can see this more in detail in the following lemma.
\begin{lemma}
	\label{pro: ori}
The control law applied in  (\ref{eq: pdynCMU}) can be implemented for each agent $i$ using only its local frame  $O_i$.
\end{lemma}

The proof is omitted due to space limitation but it is straightforward by following similar arguments as in \cite{BaArWe11,CaYuAn11}.

In order to induce some desired steady-state motion of the formation in the desired shape, we can manipulate $\mu$ and $\tilde\mu$ at the equilibrium of (\ref{eq: pdynCMU}). Then  the steady-state motion will be a function of the desired shape $z^*\in\mathcal{Z}$ and $\mu, \tilde\mu$:
\begin{align}
	\dot p^* &= \overline A(\mu,\tilde\mu)z^*.
	\label{eq: pdynZ}
\end{align}
Before discussing on how to design the motion parameters $\mu$ and $\tilde\mu$ in order to keep $\mathcal{Z}$ invariant, let us first recall some facts from rigid body mechanics \cite{Gold51}. As in the case of points in a rigid body, the steady-state velocity of every agent $\dot p^*_i$ at the desired rigid formation shape can be decomposed into
\begin{align}
\dot p_i^* &= \dot p^*_c + \underbrace{{^b\omega}\times {^bp^*_i}}_{\dot p^*_{i_\omega}},
\label{eq: motion}
\end{align}
where $^b\omega$ is the angular velocity of the rigid formation (similar to that for the rigid body). In particular, in view of (\ref{eq: pdynCMU}) with $A$ as defined in (\ref{eq: A}), the velocity (\ref{eq: motion}) at the desired shape $z^*\in\mathcal{Z}$ is given by
\begin{equation}
	\dot p^*_c + \dot p^*_{i_\omega} = \sum_{k=1}^{|\mathcal{E}|}a_{ik}z_k^*.
\label{eq: pi_motion2}
\end{equation}
Let us decompose the motion parameters into $\mu = \mu_v + \mu_\omega$ and $\tilde\mu = \tilde\mu_v + \tilde\mu_\omega$, where $\mu_v,\tilde\mu_v \in\R^{|\mathcal{E}|}$ and $\mu_\omega,\tilde\mu_\omega \in\R^{|\mathcal{E}|}$ will be used to assign the desired translational and rotational velocity, respectively, of the rigid formation. Using this decomposition, we can rewrite (\ref{eq: pi_motion2}) for all the agents into the following compact form
\begin{equation}
	\dot p^* = \underbrace{\overline A(\mu_v, \tilde\mu_v)z^*}_{\mathbf{1}_{|\mathcal{V}|\times 1}\otimes\dot p^*_c} + \underbrace{\overline A(\mu_\omega, \tilde\mu_\omega)z^*}_{\dot p^*_\omega},\label{eq: dv}
\end{equation}
where $\dot p^*_\omega \in\R^{m|\mathcal{V}|}$ is the stacked vector of all the rotational velocities $\dot p^*_{i_\omega}, i\in\{1,\dots,|\mathcal{V}|\}$.

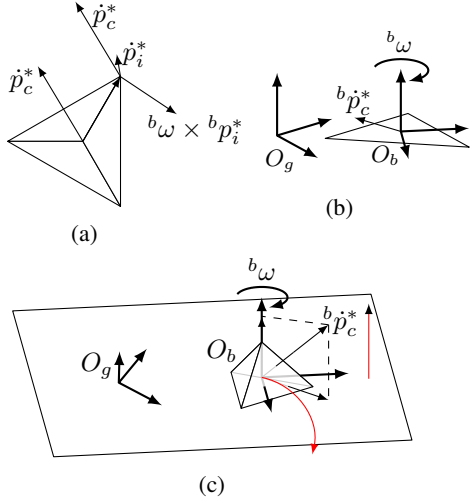
\begin{figure}
	\centering
	\begin{subfigure}{0.25\columnwidth}
\begin{tikzpicture}[line join=round]
[\tikzset{>=latex}]\filldraw[draw=black,fill=white,fill opacity=0.8](-1,0)--(0,0)--(0,0)--(.5,.866)--cycle;
\filldraw[draw=black,fill=white,fill opacity=0.8](.5,-.866)--(0,0)--(0,0)--(-1,0)--cycle;
\filldraw[draw=black,fill=white,fill opacity=0.8](.5,.866)--(0,0)--(0,0)--(.5,-.866)--cycle;
\draw[draw=black,arrows=->](0,0)--(-.6,1);
\draw[draw=black,arrows=->](.5,.86)--(-.1,1.86);
\draw[draw=black,arrows=->](.5,.86)--(1.25,.36);
\draw[draw=black,arrows=->](.5,.86)--(.45,1.16);
\draw[draw=black,arrows=->](0,0)--(.5,.86);
\node at (-.8,.8) {$\dot p^*_c$};\node at (1.5,.16) {$^b\omega \times {^bp^*_i}$};\node at (.3,1.66) {$\dot p^*_c$};\node at (.7,1.16) {$\dot p^*_i$};\end{tikzpicture}
	\caption{}
	\end{subfigure} \quad\quad\quad
	\begin{subfigure}{0.25\columnwidth}
\begin{tikzpicture}[line join=round]
[\tikzset{>=latex}]\filldraw[fill=white](0,0)--(1.916,-.116)--(1.101,.332)--cycle;
\draw[arrows=<->,thick](-.095,-.26)--(-.639,.039)--(.095,.26);
\draw[arrows=->](1.006,.093)--(.387,.277);
\draw[arrows=<->,thick](1.109,-.277)--(1.006,.093)--(1.913,.135);
\draw[arrows=->,thick](1.006,.093)--(1.006,.927);
\draw[thick](1.223,.786)--(1.228,.788)--(1.233,.789)--(1.238,.791)--(1.243,.793)--(1.248,.794)--(1.252,.796)--(1.257,.798)--(1.262,.8)--(1.266,.802)--(1.271,.803)--(1.275,.805)--(1.279,.807)--(1.283,.809)--(1.287,.811)--(1.291,.813)--(1.295,.815)--(1.299,.817)--(1.303,.819)--(1.307,.822)--(1.31,.824)--(1.314,.826)--(1.317,.828)--(1.32,.83)--(1.324,.833)--(1.327,.835)--(1.33,.837)--(1.332,.839)--(1.335,.842)--(1.338,.844)--(1.341,.847)--(1.343,.849)--(1.345,.851)--(1.348,.854)--(1.35,.856)--(1.352,.859)--(1.354,.861)--(1.356,.864)--(1.358,.866)--(1.359,.869)--(1.361,.871)--(1.362,.874)--(1.363,.876)--(1.365,.879)--(1.366,.881)--(1.367,.884)--(1.368,.886)--(1.368,.889)--(1.369,.892)--(1.37,.894)--(1.37,.897)--(1.37,.899)--(1.371,.902)--(1.371,.905)--(1.371,.907)--(1.371,.91)--(1.37,.912)--(1.37,.915)--(1.37,.918)--(1.369,.92)--(1.368,.923)--(1.368,.925)--(1.367,.928)--(1.366,.93)--(1.365,.933)--(1.363,.936)--(1.362,.938)--(1.361,.941)--(1.359,.943)--(1.357,.946)--(1.356,.948)--(1.354,.951)--(1.352,.953)--(1.35,.956)--(1.348,.958)--(1.345,.96)--(1.343,.963)--(1.34,.965)--(1.338,.968)--(1.335,.97)--(1.332,.972)--(1.329,.975)--(1.326,.977)--(1.323,.979)--(1.32,.981)--(1.317,.984)--(1.313,.986)--(1.31,.988)--(1.306,.99)--(1.303,.992)--(1.299,.994)--(1.295,.997)--(1.291,.999)--(1.287,1.001)--(1.283,1.003)--(1.279,1.005)--(1.275,1.006)--(1.27,1.008)--(1.266,1.01)--(1.261,1.012)--(1.257,1.014)--(1.252,1.016)--(1.247,1.017)--(1.243,1.019)--(1.238,1.021)--(1.233,1.022)--(1.228,1.024)--(1.223,1.026)--(1.217,1.027)--(1.212,1.029)--(1.207,1.03)--(1.202,1.031)--(1.196,1.033)--(1.191,1.034)--(1.185,1.035)--(1.18,1.037)--(1.174,1.038)--(1.168,1.039)--(1.163,1.04)--(1.157,1.041)--(1.151,1.042)--(1.145,1.043)--(1.139,1.044)--(1.133,1.045)--(1.127,1.046)--(1.121,1.047)--(1.115,1.048)--(1.109,1.049)--(1.103,1.049)--(1.097,1.05)--(1.091,1.051)--(1.084,1.051)--(1.078,1.052)--(1.072,1.052)--(1.066,1.053)--(1.059,1.053)--(1.053,1.053)--(1.047,1.054)--(1.04,1.054)--(1.034,1.054)--(1.028,1.054)--(1.021,1.055)--(1.015,1.055)--(1.009,1.055)--(1.002,1.055)--(.996,1.055)--(.989,1.055)--(.983,1.054)--(.977,1.054)--(.97,1.054)--(.964,1.054)--(.958,1.053)--(.951,1.053)--(.945,1.053)--(.939,1.052)--(.933,1.052)--(.926,1.051)--(.92,1.051)--(.914,1.05)--(.908,1.049)--(.902,1.049)--(.896,1.048)--(.889,1.047)--(.883,1.046)--(.877,1.045)--(.871,1.044)--(.866,1.043)--(.86,1.042)--(.854,1.041)--(.848,1.04)--(.842,1.039)--(.837,1.038)--(.831,1.037)--(.826,1.035)--(.82,1.034)--(.815,1.033)--(.809,1.031)--(.804,1.03)--(.798,1.029)--(.793,1.027)--(.788,1.025)--(.783,1.024)--(.778,1.022)--(.773,1.021)--(.768,1.019)--(.763,1.017)--(.759,1.016)--(.754,1.014)--(.749,1.012)--(.745,1.01)--(.74,1.008)--(.736,1.006)--(.732,1.004)--(.728,1.002)--(.724,1)--(.72,.998)--(.716,.996)--(.712,.994)--(.708,.992)--(.704,.99)--(.701,.988)--(.697,.986)--(.694,.984)--(.691,.981)--(.688,.979)--(.684,.977)--(.681,.975)--(.679,.972)--(.676,.97)--(.673,.968)--(.67,.965)--(.668,.963)--(.666,.96)--(.663,.958)--(.661,.955)--(.659,.953);
\draw[arrows=->,thick](-.639,.039)--(-.639,.873);
\draw[thick,->](1.271,.803)--(1.109,.763);
\node at (-.598,-.296) {\small $O_g$ \normalsize};\node at (.8,-.222) {\small $O_b$ \normalsize};\node at (.984,1.293) {\small $^b\omega$ \normalsize};\node at (.387,.506) {$^b\dot p^*_c$};\end{tikzpicture}
	\caption{}
	\end{subfigure}

	\begin{subfigure}{0.6\columnwidth}
\begin{tikzpicture}[line join=round]
[\tikzset{>=latex}]\filldraw[fill=white](-2.353,1.014)--(-1.808,-.927)--(2.955,-.705)--(2.41,1.236)--cycle;
\filldraw[draw=black,fill=white,fill opacity=0.8](.55,.197)--(.958,.598)--(.958,.598)--(1.638,.005)--cycle;
\filldraw[draw=black,fill=white,fill opacity=0.8](.685,-.283)--(.958,.598)--(.958,.598)--(.55,.197)--cycle;
\draw[draw=red,arrows=->](2.381,.111)--(2.381,1.112);
\draw[draw=black,arrows=<->,thick](-.368,-.243)--(-.942,.059)--(-.569,.51);
\draw[arrows=-,thick](.992,.009)--(.958,.129)--(1.468,.153);
\draw[draw=black,arrows=-](.958,.129)--(.958,.598);
\draw[draw=black,arrows=-](.958,.129)--(1.148,.279);
\draw[arrows=-,thick](.958,.129)--(.958,.598);
\draw[draw=black,arrows=-](.958,.129)--(1.309,.01);
\filldraw[draw=black,fill=white,fill opacity=0.8](1.638,.005)--(.958,.598)--(.958,.598)--(.685,-.283)--cycle;
\draw[arrows=->,thick](1.468,.153)--(2.115,.183);
\draw[draw=black,arrows=->](1.309,.01)--(1.847,-.172);
\draw[draw=black,arrows=-](1.148,.279)--(1.378,.46);
\draw[thick](1.162,1.08)--(1.167,1.082)--(1.171,1.083)--(1.176,1.085)--(1.18,1.086)--(1.185,1.088)--(1.189,1.09)--(1.194,1.091)--(1.198,1.093)--(1.202,1.095)--(1.206,1.097)--(1.211,1.098)--(1.215,1.1)--(1.218,1.102)--(1.222,1.104)--(1.226,1.106)--(1.23,1.108)--(1.233,1.11)--(1.237,1.112)--(1.24,1.114)--(1.244,1.116)--(1.247,1.118)--(1.25,1.12)--(1.253,1.122)--(1.256,1.124)--(1.259,1.126)--(1.262,1.128)--(1.264,1.13)--(1.267,1.133)--(1.27,1.135)--(1.272,1.137)--(1.274,1.139)--(1.277,1.141)--(1.279,1.144)--(1.281,1.146)--(1.283,1.148)--(1.285,1.151)--(1.286,1.153)--(1.288,1.155)--(1.29,1.158)--(1.291,1.16)--(1.292,1.162)--(1.294,1.165)--(1.295,1.167)--(1.296,1.17)--(1.297,1.172)--(1.297,1.174)--(1.298,1.177)--(1.299,1.179)--(1.299,1.182)--(1.3,1.184)--(1.3,1.187)--(1.3,1.189)--(1.3,1.191)--(1.3,1.194)--(1.3,1.196)--(1.3,1.199)--(1.3,1.201)--(1.299,1.204)--(1.299,1.206)--(1.298,1.208)--(1.297,1.211)--(1.297,1.213)--(1.296,1.216)--(1.295,1.218)--(1.293,1.22)--(1.292,1.223)--(1.291,1.225)--(1.289,1.228)--(1.288,1.23)--(1.286,1.232)--(1.284,1.235)--(1.283,1.237)--(1.281,1.239)--(1.279,1.242)--(1.277,1.244)--(1.274,1.246)--(1.272,1.248)--(1.269,1.251)--(1.267,1.253)--(1.264,1.255)--(1.262,1.257)--(1.259,1.259)--(1.256,1.261)--(1.253,1.263)--(1.25,1.266)--(1.247,1.268)--(1.243,1.27)--(1.24,1.272)--(1.237,1.274)--(1.233,1.276)--(1.23,1.278)--(1.226,1.28)--(1.222,1.281)--(1.218,1.283)--(1.214,1.285)--(1.21,1.287)--(1.206,1.289)--(1.202,1.29)--(1.198,1.292)--(1.194,1.294)--(1.189,1.296)--(1.185,1.297)--(1.18,1.299)--(1.176,1.3)--(1.171,1.302)--(1.166,1.303)--(1.162,1.305)--(1.157,1.306)--(1.152,1.308)--(1.147,1.309)--(1.142,1.31)--(1.137,1.312)--(1.132,1.313)--(1.126,1.314)--(1.121,1.315)--(1.116,1.316)--(1.111,1.318)--(1.105,1.319)--(1.1,1.32)--(1.094,1.321)--(1.089,1.322)--(1.083,1.322)--(1.078,1.323)--(1.072,1.324)--(1.066,1.325)--(1.061,1.326)--(1.055,1.326)--(1.049,1.327)--(1.044,1.328)--(1.038,1.328)--(1.032,1.329)--(1.026,1.329)--(1.02,1.33)--(1.014,1.33)--(1.008,1.331)--(1.002,1.331)--(.997,1.331)--(.991,1.332)--(.985,1.332)--(.979,1.332)--(.973,1.332)--(.967,1.332)--(.961,1.332)--(.955,1.332)--(.949,1.332)--(.943,1.332)--(.937,1.332)--(.931,1.332)--(.925,1.332)--(.919,1.331)--(.913,1.331)--(.907,1.331)--(.901,1.33)--(.895,1.33)--(.89,1.329)--(.884,1.329)--(.878,1.328)--(.872,1.328)--(.866,1.327)--(.861,1.326)--(.855,1.326)--(.849,1.325)--(.843,1.324)--(.838,1.323)--(.832,1.322)--(.827,1.322)--(.821,1.321)--(.816,1.32)--(.81,1.319)--(.805,1.317)--(.8,1.316)--(.794,1.315)--(.789,1.314)--(.784,1.313)--(.779,1.312)--(.774,1.31)--(.769,1.309)--(.764,1.308)--(.759,1.306)--(.754,1.305)--(.749,1.303)--(.745,1.302)--(.74,1.3)--(.735,1.299)--(.731,1.297)--(.726,1.295)--(.722,1.294)--(.718,1.292)--(.714,1.29)--(.709,1.289)--(.705,1.287)--(.701,1.285)--(.697,1.283)--(.694,1.281)--(.69,1.279)--(.686,1.277)--(.683,1.276)--(.679,1.274)--(.676,1.272)--(.672,1.27)--(.669,1.268)--(.666,1.265)--(.663,1.263)--(.66,1.261)--(.657,1.259)--(.654,1.257)--(.651,1.255)--(.649,1.253)--(.646,1.25)--(.644,1.248)--(.642,1.246)--(.639,1.244)--(.637,1.241)--(.635,1.239)--(.633,1.237);
\draw[draw=black,dashed](1.847,.829)--(1.847,-.172);
\draw[arrows=->,thick](-.942,.059)--(-.932,.476);
\draw[draw=black,arrows=->](.958,.598)--(.958,.942);
\draw[arrows=->,thick](.958,.598)--(.958,1.193);
\draw[arrows=<-,thick](1.09,-.342)--(.992,.009);
\draw[draw=black,arrows=->](1.378,.46)--(1.847,.829);
\draw[draw=black,dashed](1.847,.829)--(.958,.942);
\draw[thick,->](1.206,1.097)--(1.055,1.059);
\node at (-1.233,.288) {$O_g$};\node at (.428,.484) {$O_b$};\node at (.958,1.568) {$^b\omega$};\node at (2.012,.878) {$^b\dot p^*_c$};\draw[red,->] (.958,.129) to [bend left=45] (1.631,-.897);\end{tikzpicture}
	\caption{}
	\end{subfigure}
	\caption{Illustration of the steady-state motion in different frames of coordinates: (a) in $O_g$, the steady-state velocity $\dot p^*_i$ can be decomposed into translational and rotational term given by (\ref{eq: motion}); (b) in $O_b$ the steady-state velocity vector $^b\dot p^*_i$ becomes constant. For 2D formations the steady-state trajectory in $O_g$ becomes a circular closed orbit since $^b\omega$ and $^b\dot p^*_c$ are perpendicular; and (c) in general for 3D formations $^b\omega$ and $^b\dot p^*_c$ are not perpendicular, and therefore the resultant trajectory (shown in red) is helical in $O_g$.}
\label{fig: motion}
\end{figure}

In the desired shape, the stacked vector $^bz^*$ of relative positions $^bz_k^*, k\in\{1, \dots,|\mathcal{E}|\}$ in $O_b$ becomes a constant. This is due to the fact that $O_b$ follows the steady-state motion, and thus according to (\ref{eq: dv}) the velocities $^b\dot p^*_c$ and $^b\dot p^*_{i_\omega}$ also become constant. In particular, if $^b\omega = 0$, then we will have a drift of the desired formation in $O_g$; otherwise for a nonzero constant vector $^b\omega$,  the formation will follow a closed circular orbit (which is always the case in 2D formations) when $^b\omega$ is perpendicular to $^b\dot p^*_c$ , or otherwise a helical trajectory in $O_g$. Figure \ref{fig: motion} shows an illustration of these facts on the steady-state velocity of the desired formation expressed in either $O_g$ and $O_b$.

Now we are ready to show how to design the motion parameters $\mu$ and $\tilde\mu$ in (\ref{eq: dv}). Note that the steady-state motion must not distort the desired shape, i.e. the motion parameters have to be designed such that $\mathcal{Z}$ is an invariant set in  (\ref{eq: dv}). The desired steady-state velocity (\ref{eq: pdynZ}) in $O_b$ can be rewritten as
\begin{align}
	^b\dot p^* &= \overline S_1D_{^bz^*}\,\mu + \overline S_2D_{^bz^*}\,\tilde\mu = \begin{bmatrix}\overline S_1D_{^bz^*} & \overline S_2D_{^bz^*}\end{bmatrix} \begin{bmatrix}\mu \\ \tilde\mu \end{bmatrix} \nonumber \\
&= T(^bz^*)\left( \begin{bmatrix}\mu_v \\ \tilde\mu_v \end{bmatrix} + \begin{bmatrix}\mu_\omega \\ \tilde\mu_\omega\end{bmatrix}\right), \nonumber
\end{align}
where $S_1$ is constructed from the incidence matrix $B$ setting its $-1$ elements to zero and $S_2 = S_1 - B$. The motion parameters will be used to design both velocities in (\ref{eq: dv}) with respect to $O_b$ as in Figure \ref{fig: motion}c since they are constant in such a frame. Note that the knowledge about $O_b$ is only needed during the design stage for the desired steady-state velocity. During the implementation of the distributed control law, the agents only need to know about their own $O_i$'s as shown in Lemma \ref{pro: ori}.

We now  demonstrate how to design the velocity ${^b\dot p^*_c}$ employing the motion parameters. It is important to note that in an infinitesimally and minimally rigid framework, the minimum number of edges associated to a node is two in $\R^2$ or three in $\R^3$. So if $^bz^*\in\mathcal{Z}$ then the relative positions $^bz_k^*$'s associated to  agent $i$ can span the whole  $\R^{2}$ for the planar formations (or $\R^{3}$ in the 3D case). This implies that the domain of the desired velocity $^b\dot p^*_i$ is the whole space and it can be assigned by an appropriate choice of $\mu$ and $\tilde\mu$ (c.f. (\ref{eq: pi_motion2})). Since the velocity ${^b\dot p^*_c}$ is the same for all the agents, we set the following requirement for $\mu_v$ and $\tilde\mu_v$
\begin{equation}
\overline B^T (\mathbf{1}_{|\mathcal{V}|\times 1}\otimes{^b\dot p_c^*}) = \overline B^TT(^bz^*)\begin{bmatrix}\mu_v \\ \tilde\mu_v \end{bmatrix} = \mathbf{0}_{m|\mathcal{E}|\times 1},
	\label{eq: condv}
\end{equation}
which implies that $[\begin{smallmatrix}\mu_v \\ \tilde\mu_v \end{smallmatrix}] \in \operatorname{Ker}\{\overline B^TT(^bz^*)\}$. However, since in general the kernel of $T(^bz^*)$ can be non-trivial, there may exist non-zero $[\begin{smallmatrix}\mu_v \\ \tilde\mu_v \end{smallmatrix}]$ such that $^b\dot p^* = \mathbf{0}_{m|\mathcal{V}|\times 1}$, which is irrelevant for our motion control. Taking this into account, the set of motion parameters $\mu_v$ and $\tilde\mu_v$ should satisfy
\begin{equation}
\begin{bmatrix}\mu_v \\ \tilde\mu_v \end{bmatrix} \in \mathcal{U}\dfb P_{\operatorname{Ker}\{T(^bz^*)\}^\bot}\left\{\operatorname{Ker}\{\overline B^TT(^bz^*)\} \right\},
	\label{eq: spacev}
\end{equation}
where $P_\mathcal{X}$ stands for the projection over the space $\mathcal{X}$. The lower bound for the degrees of freedom of choosing the elements of $\mu_v$ and $\tilde\mu_v$ for constructing $^b\dot p^*_c$ can obviously be given by the number of motion parameters of the agent that has the least number of neighbors. In other words,
\begin{equation}
	\operatorname{dim}\{\mathcal{U}\} \geq \min_i \{|\mathcal{N}_i|\},
	\label{eq: dimv}
\end{equation}
where $\mathcal{N}_i$ is defined in Section \ref{sec: preA} and $\mathcal U$  in (\ref{eq: spacev}).
Consequently, we propose the following algorithm to compute $\mu_v$ and $\tilde\mu_v$ for given ${^b\dot p^*_c}$ and $^bz^*$.
\begin{algorithm}
	\label{al: v}
	\begin{enumerate}
	\item Choose an agent $i$ with the least number of neighbors.
	\item Assign $\mu_{v_k}$ and $\tilde\mu_{v_k}$ associated to  agent $i$ that solves
		$^b\dot p^*_c = \sum_{k=1}^{|\mathcal{E}|}a_{ik}{^bz^*_k}$.
	\item Compute a basis $\mathrm{V}$ that spans the space defined in (\ref{eq: spacev}).
\item Compute a vector $v\in\R^{2|\mathcal{E}|}$ employing the basis $\mathrm{V}$ such that the previously computed $\mu_{v_k}$'s and $\tilde\mu_{v_k}$'s are the $k$'th and $(|\mathcal{E}|+k)$'th elements of $v$. Then, the rest of elements of $v$ are necessarily the motion parameters defining $^b\dot p^*_c $ for the whole formation, i.e. $v = [\begin{smallmatrix}\mu_v \\ \tilde\mu_v \end{smallmatrix}]$.
	\end{enumerate}
\end{algorithm}

From (\ref{eq: A}) it is clear that the implementation of the motion parameters is distributed, while the computation in Algorithm \ref{al: v} is centralized since it requires the knowledge of $O_b$ and $\mathcal{Z}$. On the other hand, the motion parameters only need to be computed once since the desired speed of the motion can be modified by just rescaling $\mu_v$ and $\tilde\mu_v$. For a new agent $j$ joining the formation, its corresponding motion parameters for the translational motion can be computed in a distributed way. Consider agent $j$ joins the formation: if it asks a neighboring agent $i$ about $^i\dot p^*_c$ and $O_i$, then it becomes straightforward for agent $j$ to compute the corresponding motion parameters for the desired $^j\dot p^*_c$ by only following the second step of Algorithm \ref{al: v}.

Now we proceed to show how to design the motion parameters for the rotational velocity $^b\dot p^*_\omega$ in (\ref{eq: dv}), i.e. the rotational motion of the agents around the centroid. Note that this rotational motion preserves the constant norms $||z_k^*||$'s  but not the relative positions $z_k^*$'s. Since $\frac{1}{2}\frac{\mathrm{d}||z_k||^2}{\mathrm{d}t} = z_k^T\dot z_k$, we can impose the following condition for maintaining constant $||z_k^*||$,
\begin{align}
D_{z^*}^T {\dot z^*} = D_{z^*}^T\overline B^TT(z^*)\begin{bmatrix}\mu_\omega \\ \tilde\mu_\omega \end{bmatrix} = \mathbf{0}_{|\mathcal{E}|\times 1}.
	\label{eq: condw0}
\end{align}
As discussed after equation (\ref{eq: dv}) the relative positions $^bz^*_k$ are constant because $O_b$ is rotating with the  rigid formation. However, for having such rotation of the  formation, the constant velocities $^b\dot p_{\omega_i}^*$ must be designed in order to keep the norms $||^b z_k||$'s constant. Therefore, we impose the following restricting condition for $\mu_\omega$ and $\tilde\mu_\omega$ in addition to (\ref{eq: condw0})
\begin{align}
D_{^bz^*}^T\overline B^TT(^bz^*)\begin{bmatrix}\mu_\omega \\ \tilde\mu_\omega \end{bmatrix} = \mathbf{0}_{|\mathcal{E}|\times 1}.
	\label{eq: condw}
\end{align}
As in the case for calculating the motion parameters for $^b\dot p^*_c$, the set of motion parameters $\mu_\omega$ and $\tilde\mu_\omega$ are the subset of motion parameters such that
\begin{align}
	\begin{bmatrix}\mu_\omega \\ \tilde\mu_\omega \end{bmatrix} \in  \mathcal W\dfb P_{\mathcal{U}^\bot}\left\{
			\operatorname{Ker}\{D_{^bz^*}^T\overline B^TT(^bz^*)\}\right\},
	\label{eq: spacew}
\end{align}
where $\mathcal U$ is as in (\ref{eq: spacev}). We remark that the dimension of $\mathcal W$ in (\ref{eq: spacew}) is at least, one less than the dimension of $\mathcal U$ in (\ref{eq: spacev}). This is due to the fact $^b\dot p^*_{i_\omega}$ is perpendicular to $^bp^*_i$ (as shown in (\ref{eq: motion})) and hence, the degree-of-freedom for choosing $[\begin{smallmatrix}\mu_\omega \\ \tilde\mu_\omega\end{smallmatrix}]$ is at least one less than that for choosing $[\begin{smallmatrix}\mu_v \\ \tilde\mu_v\end{smallmatrix}]$. The calculation of the motion parameters $\mu_\omega$ and $\tilde\mu_\omega$ for a desired $^b\omega$ can be done in a similar way to Algorithm \ref{al: v}, where in the second step we assign the $a_{ik}$'s giving the constant linear velocity $^b\dot p_{i_\omega}^*$ for making the agent $i$ to rotate with angular velocity $^b\omega$ around the centroid $^bp_c^*$ derived from $\mathcal{Z}$, and the basis $\mathrm{V}$ in the third step is replaced by $\mathrm{W}$ spanning $\mathcal W$ as in (\ref{eq: spacew}).

	As before the implementation of $\mu_\omega$ and $\tilde\mu_\omega$ is fully distributed but their computation at the design stage is not. On the other hand, they only need to be computed once and the angular speed can be regulated by just rescaling their magnitude. Note that for designing different trajectories in $O_g$ as in Figure \ref{fig: motion}, we only need to change the speed of $^b\dot p^*_c$ and $^b\omega$. Unfortunately, if a new agent $j$ joins the formation, then the centroid will change, requiring to recompute in a centralized way $\mu_\omega$ and $\tilde\mu_\omega$. Nevertheless, if the \emph{instantaneous center of rotation} \cite{Gold51} $p_r\in\R^m$ is independent of the new agent, e,g, as we will see in the target enclosing problem where it will be fixed at agent $1$, then agent $j$ can still compute its motions parameters by only asking local information from a neighboring agent $i$, i.e. its $^i\dot p^*_i$ and $^i\dot p^*_c$ along with $O_i$ and in addition we also require agent $i$ to know about $^i(p_i - p_r)$.

\begin{example}
	We would like to highlight the particular example of calculating the bluemotion parameters for triangular formations and its extension to tetrahedra in 3D. Consider a generic triangular formation consisting of three agents whose $\mathbb{G}$ has the following incidence matrix
\begin{equation}
	B = \begin{bmatrix}
		1 & 0 & -1 \\
		-1 & 1 & 0 \\
		0 & -1 & 1
	\end{bmatrix}.
	\label{eq: Btrian}
\end{equation}
Then it can be checked that the following conditions for $\mu_v$ and $\tilde\mu_v$ (any four implying the remaining one)
\begin{equation}
	\left.\begin{array}{rc}
	\mu_{v_1} + \mu_{v_2} + \mu_{v_3} &= 0 \\
	\tilde\mu_{v_1} + \tilde\mu_{v_2} + \tilde\mu_{v_3} &= 0 \\
	\mu_{v_2} + \tilde\mu_{v_3} &= 0 \\
	\mu_{v_3} + \tilde\mu_{v_1} &= 0 \\
	\mu_{v_1} + \tilde\mu_{v_2} &= 0\end{array}\right\},
	\label{eq: condmuvtr}
\end{equation}
 satisfy (\ref{eq: spacev}) and they have a non-trivial solution with two degrees of freedom satisfying (\ref{eq: dimv}). We would like to emphasize that the set of equations in (\ref{eq: condmuvtr}) is independent of $\mathcal{Z}$. In general, the computation of $\mu_\omega$ and $\tilde\mu_\omega$ depends on $\mathcal{Z}$ as it is shown in the following proposition
\begin{proposition}
	\label{pro: tri}
	If ${^bz^*\in}\mathcal{Z}$ defines an equilateral triangle, then
	\begin{equation*}
	\begin{bmatrix}\mu_\omega \\ \tilde\mu_\omega \end{bmatrix} = \frac{\mathrm{w}}{\sqrt{6}}\,\textbf{1}_{6\times 1}.
	\end{equation*}
	If ${^bz^*\in}\mathcal{Z}$ defines an isosceles triangle with $d_1 = d_2$, then
	\begin{equation}
	\begin{bmatrix}\mu_\omega \\ \tilde\mu_\omega \end{bmatrix} = \mathrm{w}\alpha \sbm{\frac{3d_3^2}{2d_1^2+d_3^2} & 2-\frac{3d_3^2}{2d_1^2+d_3^2} & 1 & 2-\frac{3d_3^2}{2d_1^2+d_3^2} & \frac{3d_3^2}{2d_1^2+d_3^2} & 1}^T,
\label{eq: condiso}
	\end{equation}
	with $\alpha\in\R^+$ being a normalizing factor and $\mathrm{w}\in\R^+$ defining the angular speed.
	\label{prop: mus}
\end{proposition}

We omit the proof due to space constraint. Another interesting property is that the centroid of the triangular formation $p_c(t)$ is invariant if $\mu = \mu_\omega$ and $\tilde\mu = \tilde\mu_\omega$ belong to (\ref{eq: spacew}) since
\begin{align}
	\dot p_c(t) = \frac{1}{3} \sum_{i=1}^3\dot p_i(t) =\frac{1}{3} \sum_{k=1}^3 (\mu_{\omega_k}+\tilde\mu_{\omega_k})z_k(t) = 0. \nonumber
\end{align}
These results can be extended in a similar way to tetrahedrons.
\end{example}

\section{Stability analysis}
\label{sec: sta}
After we have the design of the desired steady-state behavior of the formation as discussed thoroughly in Section \ref{sec: motion}, we provide the stability analysis of the closed-loop system in this section. In particular we show that the control law applied in (\ref{eq: pdynCMU}), with $A$ designed as in Section \ref{sec: motion}, indeed steers the formation, at least locally, to the desired formation motion.

Similar to (\ref{eq: zdynNOmu}) and (\ref{eq: edyn}), we derive from (\ref{eq: pdynCMU}) the following closed-loop system with the motion parameters
\begin{align}
	\dot z &= -c\overline B^TR(z)^Te + \overline B^T\overline A z \label{eq: zdynM} \\
	\dot e &= -2cR(z)R(z)^Te + 2R(z)\overline A z. \label{eq: edynM}
\end{align}
Note that as shown in Proposition \ref{pro: staE}, the error system (\ref{eq: edynM}) is autonomous since the elements of $R(z)\overline Az \in\R^{|\mathcal{E}|}$ are linear combinations of the inner products $z_i^Tz_j = g_{ij}(e)$. Therefore we can write (\ref{eq: edynM}) as
\begin{equation}
	\dot e = -2cQ(e)^Te + 2f(e),
	\label{eq: edyne}
\end{equation}
with $Q(e)$ as in (\ref{eq: Qe}) with $l=2$ and $f(e) = R(z)\overline A(\mu, \tilde\mu)z$. Note that $f(e)$ is closely related to the desired shape $\mathcal{Z}$ since $z\in\mathcal{Z}$ implies $f(0) = \mathbf{0}_{|\mathcal{E}|\times 1}$. However, in general $z\in\mathcal{D}$ does not necessarily imply $f(0) = \mathbf{0}_{|\mathcal{E}|\times 1}$ since
\begin{equation*}
R(z)\overline A(\mu, \tilde\mu)z = D_z^T\overline B^TT(z)\begin{bmatrix}\mu \\ \tilde\mu \end{bmatrix}
\end{equation*}
is designed to be zero only at $z\in\mathcal{Z}$ but not at arbitrary $z\in\mathcal{D}$.
\begin{theorem}
	\label{th: thM}
There exist constants $\rho, c^*>0$ such that for system (\ref{eq: edyne}), $e=0$ corresponding to $z\in\mathcal{Z}$ with the motion parameters $[\begin{smallmatrix}\mu_v \\ \tilde\mu_v\end{smallmatrix}]$ and $[\begin{smallmatrix}\mu_\omega \\ \tilde\mu_\omega\end{smallmatrix}]$ belonging to the spaces (\ref{eq: spacev}) and (\ref{eq: spacew}) respectively is locally exponentially stable for all $c\geq c^*$ in the compact set $\mathcal{Q}\dfb\{e : ||e||^2\leq \rho\}$. In particular, the formation will converge exponentially to the shape defined by $\mathcal{Z}$ and the velocity $^b\dot p_i$ will converge exponentially to $^b\dot p^*_i$ for $i\in\{1,\dots,|\mathcal{V}|\}$.
\label{th: reconciled}
\end{theorem}
\begin{IEEEproof}
We divide the proof in two parts. Firstly we analyze the inter-agent distance dynamics and show that the shape formed eventually by the agents is the desired shape. Secondly we analyze the individual agent dynamics in order to show that they converge to the steady-state motion defined by the motion parameters.

Consider the following candidate Lyapunov function
\begin{equation*}
	V = \frac{1}{4}||e||^2,
\end{equation*}
with its time derivative  given by
\begin{align}
	\frac{\mathrm{d}V}{\mathrm{d}t} &= \frac{1}{2}e^T\dot e = -c\,e^TQ(e)e + e^Tf(e). \nonumber 
\end{align}
As shown in Proposition \ref{pro: staE}, there exists a constant $\rho > 0$ such that $Q(e)$ is positive definite in the compact set $\mathcal{Q}\dfb\{e : ||e||^2\leq \rho\}$ since $Q(0)$ is positive definite as the framework is infinitesimally and minimally rigid at $z\in\mathcal{Z}$. Since $f(e)$ is real analytic with $f(0) = \mathbf{0}_{|\mathcal{E}|\times 1}$ at $z\in\mathcal{Z}$ because of (\ref{eq: condv}) and (\ref{eq: condw}), it is also locally Lipschitz in $\mathcal{Q}$. Therefore there exists a constant $q\in\R^+$ such that
\begin{align}
	\frac{\mathrm{d}V}{\mathrm{d}t} &\leq -c \lambda_{\text{min}} ||e||^2 + e^Tf(e) \nonumber \\
&\leq  (-c\lambda_{\text{min}} + q) ||e||^2,
	\label{eq: elya}
\end{align}
where $\lambda_{\text{min}}$ is the minimum eigenvalue of $Q(e)$ in $\mathcal{Q}$. Thus if one chooses $c \geq c^* > \frac{q}{\lambda_{\text{min}}}$, then the local exponential stability of the origin of (\ref{eq: edyne}) follows, showing that the formation converges exponentially to the desired shape defined by $\mathcal{Z}$.

For the second part of the proof, we substitute $e(t)\to{\mathbf{0}_{|\mathcal{E}|\times 1}}$ and $z(t)\to\mathcal{Z}$ as $t\to\infty$ into (\ref{eq: pdynCMU}), which gives us
\begin{equation*}
	\dot p(t) - \bar A(\mu,\tilde\mu)z(t) \to 0, \, \text{as} \,\, t\to\infty.
\end{equation*}
In other words, the velocity of the agents converges exponentially to the desired velocities given by $^b\dot p^*_c$ and $^b\dot p^*_{i_\omega}$ for all $i$.
\end{IEEEproof}

\begin{example}
\label{ex: ftri}
As before, triangles and tetrahedrons can be considered as particular cases. For triangles we can write
\begin{equation*}
	f(e) = R(z)\overline Az = \frac{1}{2}F(\mu,\tilde\mu)(e+d),
\end{equation*}
where $d = \begin{bmatrix}d_1^2 & d_2^2 & d_3^2\end{bmatrix}^T$ and $F(\mu, \tilde\mu)\in\R^{3\times 3}$ is given by
\begin{align}
	F(\mu, \tilde\mu) = \left[
	\begin{smallmatrix}
	2(\mu_1-\tilde\mu_1)+\mu_2-\tilde\mu_3 & \mu_2+\tilde\mu_3 & -\mu_2-\tilde\mu_3 \\
	-\mu_3 - \tilde\mu_1 & 2(\mu_2-\tilde\mu_2)+\mu_3-\tilde\mu_1 & \mu_3+\tilde\mu_1 \\
	\mu_1+\tilde\mu_2 & -\mu_1-\tilde\mu_2 & 2(\mu_3-\tilde\mu_3)+\mu_1-\tilde\mu_2
\end{smallmatrix}\right]. \nonumber
\end{align}
We would like to pinpoint two special cases. Firstly we notice that for $\mu_\omega = 0$ and $\tilde\mu_\omega = 0$ we have that $\overline B^T\overline Az$ is zero in (\ref{eq: zdynM}) since $z_1+z_2+z_3 = 0$. In this case, $f(e) = 0$ for all $e$. Secondly, if $\mathcal{Z}$ defines an equilateral triangle, then $F$ is skew symmetric in view of  Proposition \ref{pro: tri}. This means that $e^TFe = 0$ for all $e$ in (\ref{eq: elya}). Therefore both special cases have $q = 0$, and hence for any $c>0$ the origin of (\ref{eq: edynM}) is locally exponentially stable. As before, similar results can be easily extended to tetrahedrons.
\end{example}

It is important to note that we are assuming that the measurements done by the agents are not biased with respect to each other, i.e. we do not consider any \emph{undesired} mismatches that will affect the performance of the motion controller. This is still an open issue that could be addressed in future works by exploiting the ideas contained in \cite{MarCaoJa15}.

In the following section, we apply the motion parameters to a more specific applications of formation motion control.

\section{Applications of employing only motion parameters}

The results from Theorem \ref{th: thM} allow one to design the behaviour of the steady-state motion of the desired formation. However, these results are restricted to the design of the velocities $^b\dot p^*_i$ for all $i$, which are obviously defined in $O_b$. In particular, for the case of having only translational motion, i.e. $[\begin{smallmatrix}\mu_\omega \\ \tilde\mu_{\omega}\end{smallmatrix}] = \mathbf{0}_{2|\mathcal{E}|\times 1}$, the steady-state orientation of $O_b$ with respect to $O_g$ is unknown and depends on the initial condition $p(0)$. Similarly, for the rotational motion around the centroid, i.e. $[\begin{smallmatrix}\mu_v \\ \tilde\mu_v\end{smallmatrix}] = \mathbf{0}_{2|\mathcal{E}|\times 1}$, the steady-state position of the centroid $p_c$ of the formation also depends on $p(0)$.

	Throughout the following two subsections, we are going to extend the results of Theorem \ref{th: thM} in order to overcome the issues mentioned above. The first extension is concerned with the control of the steady-state orientation of $O_b$ with respect to $O_g$ in the special case of having only a translational motion. The second extension deals with the control of the instantaneous center of rotation for the formation, which is closely related to the problem of tracking and enclosing a target as briefly mentioned in the Introduction.

\subsection{Translational motion with controlled heading}
\label{sec: tra}
Suppose that the motion parameters have been assigned such that the steady-state motion of the formation defines a pure translation, i.e. $[\begin{smallmatrix}\mu_\omega \\ \tilde\mu_{\omega}\end{smallmatrix}] = \mathbf{0}_{2|\mathcal{E}|\times 1}$. In this special case, we are interested in controlling the steady-state orientation of $O_b$ with respect to $O_g$, and consequently the heading of the translational motion in $O_g$. This task can be accomplished if we control only one relative position among all the $^bz_k$'s with respect to $O_g$. Without loss of generality we set the relative position $z_1$ associated to  edge $\mathcal{E}_1 = (1, 2)$ for achieving this task. Since $z^*\in\mathcal{Z}$ defines a rigid formation, if we control the orientation of $z_1$ then the orientation of the rest $z_k$'s will also be  controlled. In other words, we are setting the desired orientation of the formation in $O_g$ with the following relation
\begin{equation}
	z_1^* = {_b^gR}\,{^bz_1^*},
	\label{eq: bgR}
\end{equation}
where $z_1^*$ has the orientation with respect to $O_g$ given by the desired rotational matrix $_b^gR\in\R^{m\times m}$ defining the steady-state orientation of $O_b$ with respect to $O_g$.
	
Let us introduce the following potential function in order to settle the orientation controller for $z_1$ implemented by the agents in $\mathcal{E}_1$
\begin{equation}
	V_{1o} = \frac{1}{2}||z_1 - z_1^*||^2,
	\label{eq: Vor}
\end{equation}
and define the alignment error 
\begin{equation*}
e_{1o} = z_1-z_1^*.
\end{equation*}
The orientation controller derived from the gradient  of (\ref{eq: Vor}) is
\begin{equation}
	\nabla_{p_1} V_{1o} = -\nabla_{p_2} V_{1o} = e_{1o}. \label{eq: e1o}
\end{equation}
Note that we require  agents $1$ and $2$ to know $O_g$, since they will need $^iz_1^*$ for the computation of (\ref{eq: e1o}) in $O_i$.
Let us introduce some definitions before writing in a compact form the dynamics of the agents including the controller (\ref{eq: e1o}). Define the orientation error vector
\begin{equation*}
e_o \dfb \operatorname{col}\{e_{1o}, \dots, e_{ko}\}\in\R^{m|\mathcal{E}|},
\end{equation*}
where $e_{ko} = \mathbf{0}_{m\times 1}$ for all $k\neq1$, and  the following \emph{augmented} incidence matrix
\begin{align}
	B_a &= \begin{bmatrix}B_o & \mathbf{0}_{2\times (|\mathcal{E}|-1)}  \\
	\mathbf{0}_{(|\mathcal{V}|-2)\times 1} &\mathbf{0}_{(|\mathcal{V}|-2)\times (|\mathcal{E}|-1)}
\end{bmatrix}\in\R^{|\mathcal{V}|\times |\mathcal{E}|},\nonumber
\end{align}
where 
\begin{equation*}
B_o = \begin{bmatrix}1 & -1\end{bmatrix}^T,
\end{equation*}
describes the neighbor relationships for the alignment task and $B_a$ has been adjusted to have the same dimensions of $B$. Now we can extend the control law in (\ref{eq: pdynCMU}) including the controller (\ref{eq: e1o}) in agents $1$ and $2$ as follows
\begin{equation}
	u = c (-R(z)^Te -\overline B_ae_o) + \overline A z.
	\label{eq: controlOr}
\end{equation}
As we have done earlier, we derive the following closed-loop system by substituting (\ref{eq: controlOr}) into (\ref{eq: p})
\begin{align}
	\dot z &= -c(\overline B^TR(z)^Te -\overline B^T\overline B_ae_o) + \overline B^T \overline A z \nonumber \\
	\dot e &= -2c(R(z)R(z)^Te - R(z)^T\overline B_ae_o) + 2D_z^T\overline B^T \overline A z \label{eq: eddyn} \\
	\dot e_o &= \overline B^T_a \dot p = -c(\overline B^T_aR(z)^Te -\overline B^T_a\overline B_ae_o) + \overline B^T_a \overline A z  . \label{eq: eodyn}
\end{align}
Without restricting the value for the desired velocity $^b\dot p_c^*$, we introduce the following assumptions on $\mathbb G$ and the construction of $^b\dot p_c^*$.
\begin{assumption}
	\label{as: t}
	For 2D (3D) formations the subset of agents $\mathcal{H}\dfb\{1, 2, 3, (4)\}$ forms a complete subgraph $\mathbb G_h \subseteq \mathbb G$. In addition for the desired shape $z^*\in\mathcal{Z}$ the agents in $\mathcal{H}$ form a non-degenerated triangle (tetrahedron). The desired velocities $^b\dot p_i^* = {^b\dot p_c^*}$ for $i\in\mathcal{H}$ are constructed employing the motion parameters derived from \emph{only} $\mathbb G_h$.
\end{assumption}

Under Assumption \ref{as: t}, from Example \ref{ex: ftri} we have that
\begin{equation}
	\overline B^T_a \overline A z = \mathbf{0}_{m|\mathcal{E}|\times 1}.
	\label{eq: as}
\end{equation}
The Assumption \ref{as: t} is imposed only on the agents in $\mathcal{H}$ in order to make easier the following analysis. For the other agents we construct their motion parameters $\mu_v$ and $\tilde\mu_v$ with the only condition of satisfying (\ref{eq: spacev}).

We are ready now to present and prove the following convergence result.
\begin{theorem}
	\label{th: ori}
Suppose that $A$ as in (\ref{eq: A}) is not zero and its motion parameters $[\begin{smallmatrix}\mu \\ \tilde\mu\end{smallmatrix}]$ define a pure translational motion and Assumption \ref{as: t} is satisfied. Then, there exists constants $\rho,c^* > 0$ such that the origin of the error systems (\ref{eq: eddyn}) and (\ref{eq: eodyn}) are locally exponentially stable for all $c\geq c^*$ in the compact set $\mathcal{Q}\dfb\{e, e_{1o}:||\begin{smallmatrix}e \\ e_{1o}\end{smallmatrix}||^2\leq\rho\}$. In particular, the formation will converge exponentially to the shape defined by $z\in\mathcal{Z}$ with $z_1 = z_1^*$, and the velocity of the centroid $\dot p_c$ converges exponentially to $_b^gR{^b\dot p_c^*}$, where $_b^gR$ is the desired rotational matrix between $O_b$ and $O_g$ as given in (\ref{eq: bgR}).
\end{theorem}
\begin{IEEEproof}
	First we prove the convergence of the formation to the desired shape $\mathcal{Z}$. Consider the following candidate Lyapunov function
\begin{equation}
	V = \frac{1}{4}||e||^2 + \frac{1}{2}||e_o||^2,
	\label{eq: VT}
\end{equation}
whose time derivative is
\begin{align}
	\frac{\mathrm{d}V}{\mathrm{d}t} &= \frac{1}{2}e^T\dot e + e_o^T\dot e_o \nonumber = -c\Big(e^TR(z)R(z)^Te + e^TR(z)\overline B_ae_o \nonumber \\ 
	 &+e_o^T\overline B_a^TR(z)^Te +e_o^T\overline B_a^T\overline B_ae_o\Big) + e^Tf(e) \nonumber \\
	 &= -c||R(z)^Te + \overline B_a e_o||^2 + e^Tf(e),
	\label{eq: Vdd}
\end{align}
where $f(e)$ is as in (\ref{eq: edyne}) and we have employed (\ref{eq: as}) under the Assumption \ref{as: t}. Because of the zero elements of $e_o$ and $B_a$, we can further rewrite (\ref{eq: Vdd}) as
\begin{align}
\frac{\mathrm{d}V}{\mathrm{d}t} &= -c \begin{bmatrix}e^T & e_{1o}^T\end{bmatrix}Q\begin{bmatrix}e^T & e_{1o}^T\end{bmatrix}^T + e^Tf(e),
	\label{eq: Vdd2}
\end{align}
where 
\begin{equation*}
	Q = 
\begin{bmatrix}
	R(z)R(z)^T & R_{2m}(z) \overline B_o \\
	\overline B_o^TR_{2m}(z)^T & \overline B_o^T\overline B_o
\end{bmatrix},
\end{equation*}
with the matrix $R_{2m}(z)$ being the first $2m$ columns of $R(z)$, corresponding to the relative positions involving the agents $1$ and $2$. According to Theorem \ref{th: V}, the first term of (\ref{eq: Vdd2}) coming from the gradient  of a potential function is
strictly negative in the compact set $\mathcal{Q}$ for some positive constant $\rho$ excluding the origins of $e$ and $e_{1o}$. Thus $Q$ is positive definite in $\mathcal{Q}$. Since $f(e)$ is locally Lipschitz with $f(0) = \mathbf{0}_{|\mathcal{E}|\times 1}$ at $z\in\mathcal{Z}$ there exists a constant $q\in\R^+$ such that
\begin{equation*}
	\frac{\mathrm{d}V}{\mathrm{d}t}\leq (-c\lambda_{\text{min}} + q)\,\bigg |\bigg |\begin{bmatrix}e \\ e
	_{1o}\end{bmatrix} \bigg |\bigg |^2,
\end{equation*}
where $\lambda_{\text{min}}$ is the minimum eigenvalue of $Q$ in $\mathcal{Q}$.
Therefore choosing $c\geq c^* > \frac{q}{\lambda_{\text{min}}}$ implies that (\ref{eq: VT}) is not increasing if $e(0), e_{1o}(0) \in\mathcal{Q}$. Hence the origin of the system formed by (\ref{eq: eddyn}) and (\ref{eq: eodyn}) is exponentially stable.

For the second part of the proof, note that $e_{1o}(t) \to 0$ as $t\to\infty$ implies that $z_1(t)\to z_1^*$ as $t\to\infty$, and because $z(t)\to\mathcal{Z}$ as $t\to\infty$,  we have that
\begin{equation*}
	\dot p(t) - \left(\mathbf{1}_{|\mathcal{V}|\times 1} \otimes\, {_b^g}R{^b\dot p_c^*}\right) 
	\to 0, \,\text{as}\, t\to\infty,
\end{equation*}
i.e., all agents converge to the same velocity $\dot p^*_i = {_b^g}R^b\dot p_c$ .
\end{IEEEproof}
\begin{remark}
If one does not want any steady-state movement, i.e. $A=\mathbf{0}_{|\mathcal{V}|\times |\mathcal{E}|}$, then the local exponential stability of the desired aligned formation is guaranteed for any $c>0$.
\end{remark}
\begin{remark}
The constant $q$ in the proof of Theorem \ref{th: ori} is the same as that in the proof of Theorem \ref{th: thM}. For triangular and tetrahedron formations, the convergence to the desired shape with the desired heading occurs for any $c>0$ since $q=0$.
\end{remark}

It seems reasonable to assign only one agent as a leader for the orientation of the formation. In such a case the neighbor relationships for the alignment are described by a directed graph. In fact this directed graph consists of only two nodes and therefore it is strongly connected. The stability analysis can be worked out employing as a potential function the one shown in \cite{Cho12} for strongly connected directed graphs in substitution of the potential $\frac{1}{2}||e_o||^2$ in (\ref{eq: VT}).

Note that the standard approach for the translational motion of a rigid formation is to assign a velocity to a leader and the other agents are equipped with local estimators in order to track the velocity of the leader \cite{BaArWe11}. In the special case of having the formation travelling at a constant speed, we have shown in Theorem \ref{th: ori} that such estimators are not necessary, reducing the complexity of the implementation of the control law. In addition in our algorithm the order of the agents in the formation is preserved with respect to the direction of motion. Finally, in Theorem \ref{th: ori} we have shown that the convergence to the desired velocity is exponential and not only asymptotic as in the leader-follower case.

\subsection{Target enclosing problem}
\label{sec: enc}
The target enclosing problem addresses the scenario of having a team of \emph{enclosing} agents with the objective of surrounding some specific independent target which is either stationary or moving with an unknown velocity. Furthermore, in orbiting missions, we not only require the agents to surround the target but to circumnavigating it. Many work has been done in this area such as \cite{Shames12,Guo10} among others. However they usually do not allow the manipulation of the desired formation consisting of the enclosing agents together with the target; all of the agents are usually restricted to follow the same circular trajectory around the target, which is at the center of the circle. It is worth noting that in other works, such as in \cite{Mar12}, the complexity of the controller is much higher than the one proposed in this paper using the motion parameters. Furthermore, other drawbacks that are usually present in many works are that all the enclosing agents have to measure their relative positions or distances to the target, and the proposed algorithms have only been tested in 2D scenarios.

It has been shown in Section \ref{sec: motion} that the instantaneous center of rotation $p_r$ for the desired steady-state motion of the formation is fixed in $O_g$. Note that for $\dot p_c^* = \mathbf{0}_{m\times 1}$, i.e. $[\begin{smallmatrix}\mu_v \\ \tilde\mu_v\end{smallmatrix}] = \mathbf{0}_{2|\mathcal{E}|\times 1}$, the instantaneous center of rotation $p_r$ is coincident with $p_c$. However the steady-state of $p_r$ in Theorem \ref{th: thM} depends on the initial condition $p(0)$. In this section we are going to associate $p_r$ to the target to be enclosed, and without loss of generality we set $p_1 = p_r$. Furthermore, we will assume that $\dot p_1 = \hat v_1 \in \R^m$ is constant and unknown to the enclosing agents. Let us define
\begin{equation*}
B_d \dfb \tilde IB,
\end{equation*}
where $\tilde I_{|\mathcal{V}|\times |\mathcal{V}|}$ is obtained from the identity matrix setting its first element  to zero. With this operation we are setting all the elements of the first row of $B_d$ to zero. The interpretation for $B_d$ is that some of the enclosing agents are measuring their relative positions with respect to the target, and that obviously the target is not interacting with the rest of the formation.

In order to estimate the unknown $\hat v_1$ by the enclosing agents, we will use an estimator $\hat v_i$ for each enclosing agent, whose dynamics are
\begin{equation}
	\dot{\hat v}_i = -\kappa \nabla_{p_i}V,\, i\in\{2, \dots, |\mathcal{V}|\},
	\label{eq: est}
\end{equation}
where $\kappa \in\R^+$ is a constant gain and $V$ the potential function as used in the formation control. The estimator (\ref{eq: est}) can be written for all the agents in the following compact form
\begin{equation}
\dot{\hat v} = -\kappa \overline B_dD_ze,
\label{eq: esti}
\end{equation}
where $\hat v\in\R^{m|\mathcal{V}|}$ is the stacked column vector of $\hat v_i$'s. Note that $\dot{\hat v}_1 = \mathbf{0}_{m\times 1}$ corresponds to the target, i.e. it does not have any estimator.

Consider the following control law
\begin{align}
	u = -c\overline B_dD_ze  + \overline A(\mu, \tilde\mu)z +\hat v,
	\label{eq: penc}
\end{align}
where $c\in\R^+$ is a positive gain. Since all the elements of the first row of $B_d$ are zero, the first term of the control law in (\ref{eq: penc}) is not playing a role in the target's position $p_1$. The motion parameters of $A(\mu, \tilde\mu)$ have been designed to have $^bp_1 = {^bp_r}$ as the desired steady-state instantaneous center of rotation for the desired shape $z\in\mathcal{Z}$. Recall that the desired infinitesimally and minimally rigid formation includes $p_1$ and this is why we can calculate the needed motion parameters. For example if by design we set $^bp_1 = {^bp_r} = {^bp_c}$ then $[\begin{smallmatrix}\mu_v \\ \tilde\mu_v\end{smallmatrix}] = \mathbf{0}_{2|\mathcal{E}|\times 1}$ and $[\begin{smallmatrix}\mu_\omega \\ \tilde\mu_\omega\end{smallmatrix}]$ is as shown in Section \ref{sec: motion}. In general, however, the instantaneous center of rotation can be inside or outside of the area or volume defined by the enclosing agents. Note that since $p_1 = p_r$, the motion parameters of the target are zero, i.e. the elements of the first row of $A(\mu, \tilde\mu)$ are zero. Therefore the second term in  (\ref{eq: penc}) is not playing any role in the dynamics of $p_1$. This is an obvious requirement since the target is not collaborating with the rest of the enclosing agents.

	Our approach has two obvious limitations. Firstly we require to have more than two (three) agents including the target in 2D (3D) in order to form an infinitesimally and minimally rigid framework. Secondly, because of the definition of infinitesimally rigid framework, the desired shape including the target must exclude the ones where all the agents are collinear or coplanar in 2D or 3D respectively. 

Define the error velocity as
\begin{equation*}
	e_v = \hat v - (\mathbf{1}_{|\mathcal{V}|\times 1} \otimes \hat v_1).
\end{equation*}
Note that the first $m$ elements of $e_v$ are zero, therefore
\begin{equation}
\overline B^T e_v = \overline B_d^T e_v,
\label{eq: Be}
\end{equation}
and in addition one can check the following identity as well
\begin{align}
B^TB_d = B^T\tilde IB =  B^T\tilde I^T \tilde IB = B_d^TB_d.
\label{eq: BdBd}
\end{align}

Before stating the main result of this section, let us write down the closed-loop dynamics by substituting (\ref{eq: penc}) in (\ref{eq: p})
\begin{align}
	\dot z &= \overline B^T\dot p = -c\overline B_d^T\overline B_dD_ze + \overline B^TA(\mu, \tilde\mu)z + \overline B_d^Te_v \label{eq: zdynenc}\\
	\dot e &= 2D_z^T\dot z = -2cD_z^T\overline B_d^T\overline B_dD_ze  + 2f(e) + 2D_z^T\overline B_d^Te_v
	\label{eq: edynenc} \\
	\dot e_v &= \dot{\hat v} = \kappa \overline B_dD_ze, \label{eq: evdyn}
\end{align}
where we have employed the identities (\ref{eq: Be}) and (\ref{eq: BdBd}), $f(e)$ is the same as in (\ref{eq: edyne}), and for the third term of (\ref{eq: zdynenc}) we have used the following well known fact
\begin{equation}
	\overline B^T (\mathbf{1}_{|\mathcal{V}|\times 1} \otimes \hat v_1) = \overline B_d^T (\mathbf{1}_{|\mathcal{V}|\times 1} \otimes \hat v_1) = \mathbf{0}_{m|\mathcal{E}|\times 1}.
\end{equation}
\begin{theorem}
	\label{th: enc}
	Suppose that the steady-state motion of the rigid formation $^bz^*\in\mathcal{Z}$ defined by the motion parameters $\mu$ and $\tilde\mu$ has $^bp_1^*$ as the instantaneous center of rotation. Suppose that  agent $1$ is a free target with constant velocity $\hat v_1$ which is unknown to the rest of the enclosing agents. Then there exist constants $\rho, c^* >0$ such that the origins of 
(\ref{eq: edynenc}) and (\ref{eq: evdyn}) are locally asymptotically stable in the compact set $\mathcal{Q}\dfb\{e, e_v:||\begin{smallmatrix}e \\e_v\end{smallmatrix}||^2\leq\rho\}$. In particular, the formation will converge asymptotically to the shape defined by $\mathcal{Z}$, and the motion of the enclosing agents converge to a rotation around the target with angular speed $||\omega|| = \frac{||^b\dot p_i^*||}{||^bp_1^* - {^bp_i^*}||}, \, i\in\{2, \dots, |\mathcal{V}|\}$.
\end{theorem}
\begin{IEEEproof}
Consider the following candidate Lyapunov function
\begin{equation*}
	V = \frac{\kappa}{4}||e||^2 + \frac{1}{2}||e_v||^2,
\end{equation*}
where $\kappa$ is as in (\ref{eq: est}), with time derivative given by
\begin{align}
	\frac{\mathrm{d}V}{\mathrm{d}t} &= \frac{\kappa}{2}e^T\dot e + e_v^T\dot e_v \nonumber \\
									   &=-\kappa ce^TD_z^T\overline B_d^T\overline B_dD_ze + \kappa e^Tf(e) + \kappa e^TD_z^T\overline B_d^Te_v \nonumber \\ &- \kappa
	e_v^T\overline B_dD_ze \nonumber \\
	&\leq \kappa (-c\lambda_{\text{min}} + q)||e||^2,
	\label{eq: Vdenc}
\end{align}
where $f(e)$ is as in (\ref{eq: edyne}) and $q$ is obtained from $f(e)$ since it is locally Lipschitz with $f(0) = \mathbf{0}_{|\mathcal{E}|\times 1}$ if $z\in\mathcal{Z}$.
	Note that $M(e)=D_z^T\overline B_d^T\overline B_dD_z$ is not defined by the rigidity matrix $R(z)$ because of $B_d$. Nevertheless it can be checked that $M(0)$ is still positive definite if for $z\in\mathcal{Z}$ the framework is infinitesimally and minimally rigid. Therefore $M(e)$ is positive definite for some $\rho \leq \rho_1\in\mathbb{R}^+$ in the compact set $\mathcal{Q}$, and then $\lambda_{\text{min}}$ in (\ref{eq: Vdenc}) is the minimum eigenvalue of $M(e)$ in $\mathcal{Q}$. If one chooses $c \geq c^* > \frac{q}{\lambda_{\text{min}}}$, then the time derivative (\ref{eq: Vdenc}) is not increasing, and it is zero if and only if $e = 0$. For such a case, we can conclude that the shape of the formation converges to $z^*\in\mathcal{Z}$, which means that the steady-state motion of the agents will not distort the desired shape. Thus following a similar argument as in Section \ref{sec: motion}, the largest invariant set of the closed-loop system (\ref{eq: zdynenc}), (\ref{eq: edynenc}) and (\ref{eq: evdyn}) with $e = 0$ and $z^*\in\mathcal{Z}$ is given by
\begin{equation}
	\mathcal{T} \dfb \left\{\hat v\, : \, D^T_{z^*}\overline B^T\left(\overline S_1 D_{z^*}\mu + \overline S_2 D_{z^*}\tilde\mu + \hat v\right) = 0\right\}. \label{eq: vhatT}
\end{equation}
Since $\hat v_1$ in $\hat v$ is fixed, only two steady-state motions for the agents are possible when $\hat v \in \mathcal{T} = \mathcal{T}_u\cup\mathcal{T}_d$. In the set $\mathcal{T}_u$ all the agents travel with the same velocity $\dot p_i = \hat v_1$ and hence $||e_v||^2 = \rho_2 \in \mathbb{R}^+$. In the set $\mathcal{T}_d$ the enclosing agents rotate around $p_1$ with $||e_v|| = 0$. Therefore invoking the LaSalle's invariance principle, we have that the origins of $e_v$ and $e$ defining $z\in\mathcal{Z}$ are locally asymptotically stable in $\mathcal{Q}$ with $\rho < \min\{\rho_1, \rho_2\}$ and
\begin{equation}
	\dot p(t) - \overline A(\mu, \tilde\mu)z(t) - (\mathbf{1}_{|\mathcal{V}|\times 1} \otimes {\hat v_1}) \to 0, \, t\to\infty,
	\label{eq: lala}
\end{equation}
which defines the steady-state rotational movement of the enclosing agents around the target. The steady-state angular speed of the enclosing agent $i$ around the target is its speed over the radius of the rotation. Since the target is moving with velocity $\hat v_1$ we have to compensate it in the following formula
\begin{align}
	||\omega|| = \frac{||\dot p^*_i - \hat v_1||}{||p^*_1 - {p^*_i}||} 
			   = \frac{||^b\dot p_i^*||}{||^bp^*_1 - {^bp^*_i}||},
	\, i\in\{2, \dots, |\mathcal{V}|\}, \label{eq: ome}
\end{align}
where we have employed (\ref{eq: lala}).
\end{IEEEproof}
\begin{remark}
	For the triangular formation consisting of three agents, one being the target and the other two the enclosing ones, if $\mathcal{Z}$ defines an equilateral triangle, then we have that $e^Tf(e) = e^TFe = 0$ since $F$ is skew symmetric in view of Proposition \ref{pro: tri}. Therefore one can choose any $c>0$ since $q = 0$ in (\ref{eq: Vdenc}). Also note that $q$ is the same constant in the proofs of Theorems \ref{th: enc} and \ref{th: thM}.
\end{remark}
\begin{remark}
The minimum number of enclosing agents sensing their relative positions with respect to the target is the same as the minimum number of edges needed for a node belonging to an infinitesimally and minimally rigid framework. Therefore for planar frameworks the number is two and for 3D frameworks the number is three.
\end{remark}
\begin{remark}
	The estimation of the constant velocity of the target is based on the internal model principle as it has been used in \cite{BaArWe11}. This implies that the velocity of the target might be also driven by a superposition of finite sinusoidal signals with known frequency but unknown phase and amplitude. In such a case, the dynamics of the estimator (\ref{eq: esti}) have to be modified in order to include such frequencies as it has been shown in \cite{BaArWe11} or \cite{MarCaoJa15}.
\end{remark}
\begin{remark}
The enclosing agents employ their own local coordinate system in view of Lemma $\ref{pro: ori}$. Therefore each enclosing agent $i$ estimates the velocity of the target $\hat v_1$ in $O_i$.
\end{remark}

Since $\hat v_1$ is unknown, the conservative region of attraction $\mathcal{Q}$ is difficult to establish or estimate. Nevertheless we can avoid the undesired equilibrium set $\mathcal{T}_u\subset\mathcal{T}$ as in (\ref{eq: vhatT}) when $\rho_2 \leq \rho \leq \rho_1$. Firstly note that the constants $\rho_1$ and $\rho_2$ are mainly determined by $p(0)$ and $\hat v(0)$ respectively. Secondly we know that $\mathcal{T}_u$ determines a pure translation for all the agents. Suppose that the enclosing agents can measure the angular velocity $^i\omega$, e.g. with a gyroscope. If $\hat v(t)$ converges to $\mathcal{T}_u$, then $^i\omega$ will converge to zero as fast as $z(t)$ approaches $\mathcal{Z}$. Furthermore, $u_i$ for all $i$ will converge to $\hat v_1$. Therefore, we can program the agents such that when $^i\omega$ is \emph{almost zero} at $t=t^*$, their estimators are reset to $\hat v_i(t^*) = u_i(t^*)$. Since $z(t^*)$ is \emph{very close} to $\mathcal{Z}$, then certainly the errors $e(t^*)$, $e_v(t^*) \in\mathcal{Q}$ with $\rho < \rho_2$.

In the case that we do not require the enclosing agents to rotate around the target, we can combine the results of the Theorems \ref{th: ori} and \ref{th: enc}. The result of such combination is: the enclosing agents estimate the unknown velocity of the target; the enclosing agents form a rigid formation together with the target, whose orientation with respect to the target can be determined by a leader. Furthermore in such a special case we have that $A = \mathbf{0}_{|\mathcal{V}|\times|\mathcal{E}|}$, and therefore $q = 0$ in Theorems \ref{th: ori} and \ref{th: enc}.

We finish by emphasizing again that the results presented here do not need  measuring any relative velocity, and not all of the enclosing agents need to measure their relative positions with respect to the target agent.

\section{Experimental results}
\label{sec: exp}
\begin{figure}
\centering
\includegraphics[width=0.5\textwidth]{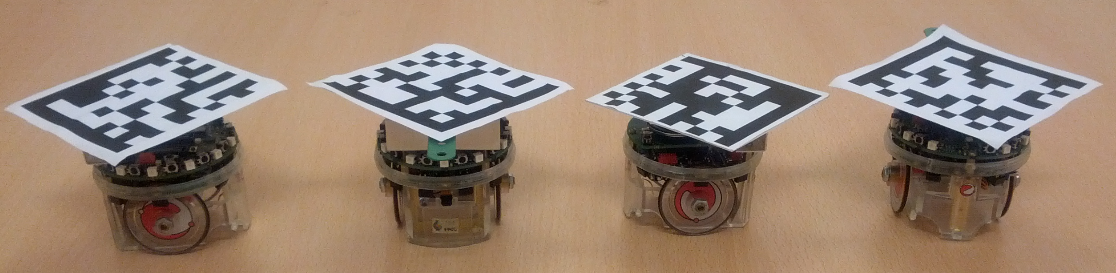}
\caption{Four wheeled E-pucks with data-matrices as markers on their tops. The code in each data-matrix corresponds to the node number for the robot in the sensing graph.}
\label{fig: epucks}
\end{figure}

In this section we validate and test the performances of the results in Theorems \ref{th: ori} and \ref{th: enc} using E-puck mobile robots \cite{epucks}. The experimental setup consists of four wheeled E-puck robots in a 2D area of about $2.6\times 2$ meters. Each robot is tagged by a data-matrix on its top as shown in Figure \ref{fig: epucks}. Since E-pucks are unicycles, we apply feedback linearization about a reference point to obtain the single-integrator dynamics in (\ref{eq: p}). In practice we control the formation of the reference points of the robots. We set as a reference point to the intersection of the two solid bars in the data matrix. We simulate the robots' on-board sensors for the relative positions with a PC. The computer runs a real time process for the computation of the relative positions between the robots with a vision algorithm, and sends the data back to the robots at a frequency of about $20$Hz via Bluetooth. The robots then apply the corresponding distributed algorithm. The whole image of the testing area is covered by $1280\times 720$ pixels, where the distance between two consecutive horizontal or vertical pixels corresponds to approximately to $2.0$mm. The raw footage 
of the experiments reported in this paper can be found in the playlist \emph{Robotics} at \url{https://www.youtube.com/c/HectorGarciadeMarina} .

\subsection{Translational motion with controlled heading experiment}
In this experiment we will validate the results from Theorem \ref{th: ori}.
We consider four robots that have a square with $225$ pixels ($45.0$mm) for the length of the side as a prescribed shape $\mathcal{Z}$. The incidence matrix describing the sensing topology for maintaining the prescribed shape and the derived $A(\mu, \tilde\mu)$ are

\begin{figure}
\centering
\begin{subfigure}{0.49\columnwidth}
\includegraphics[width=1\columnwidth]{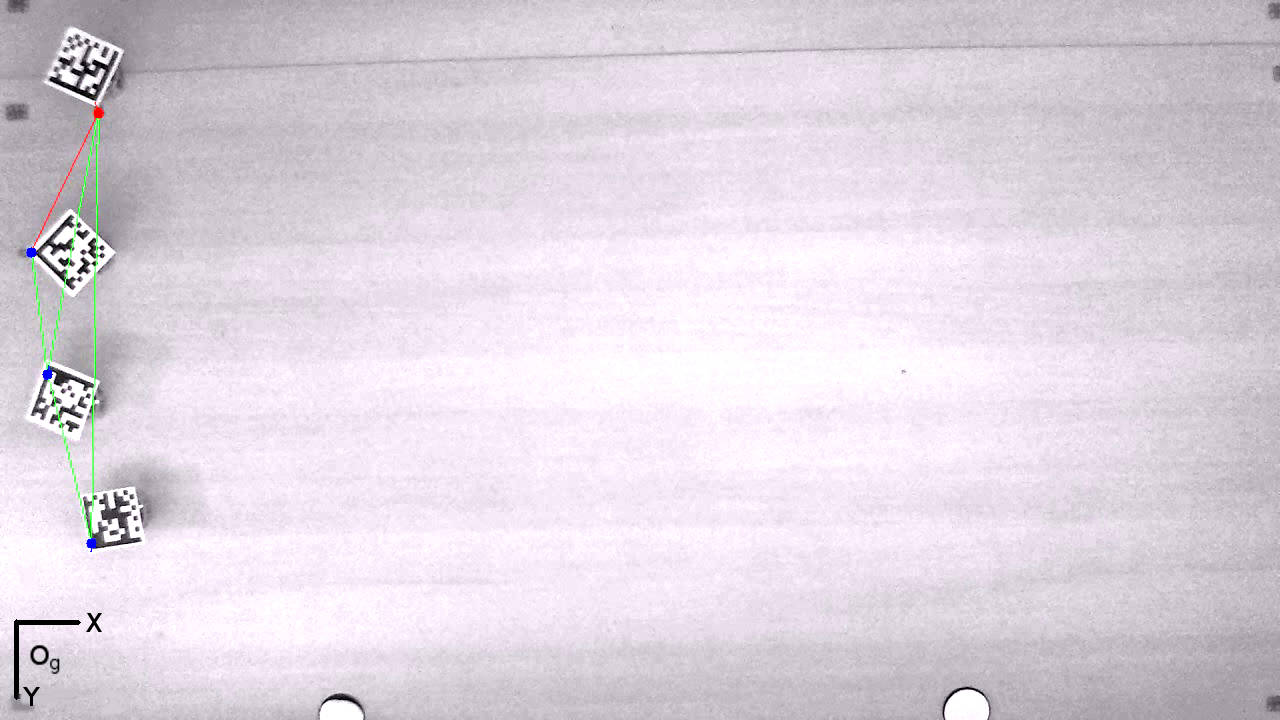}
\caption{Time $t = 0$ secs.}
\end{subfigure}
\begin{subfigure}{0.49\columnwidth}
\includegraphics[width=1\columnwidth]{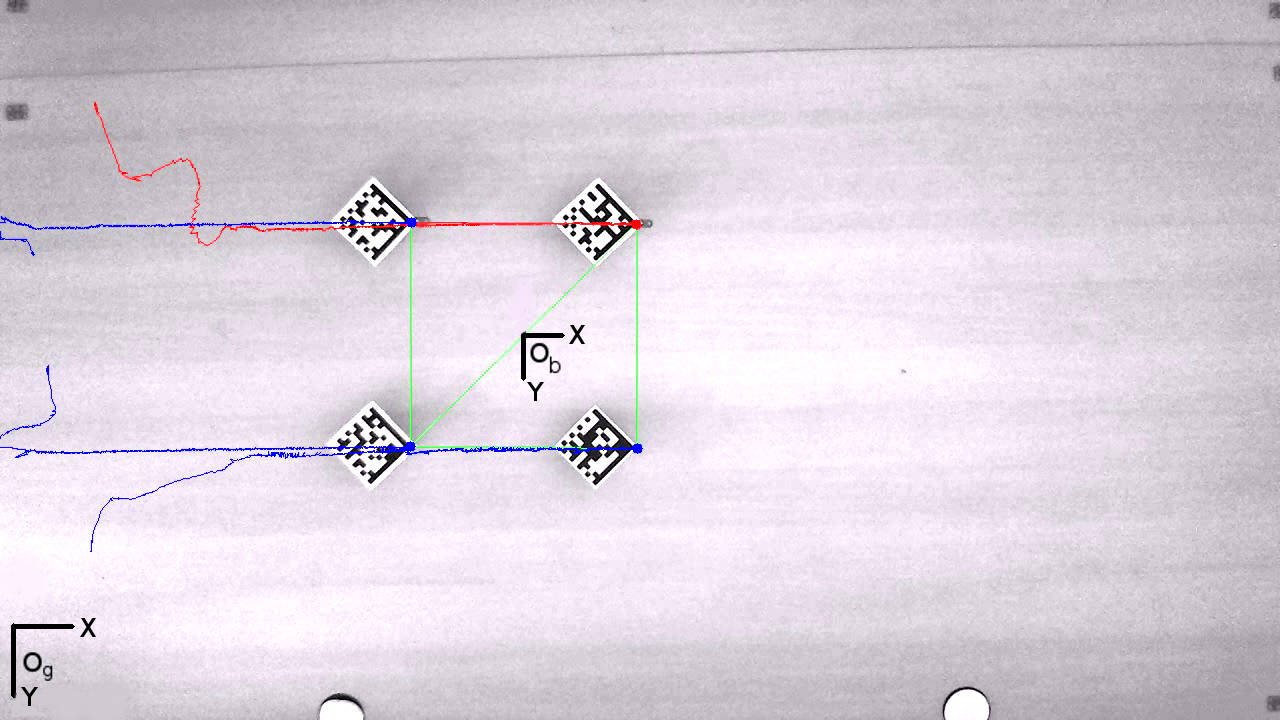}
\caption{Time $t = 90$ secs.}
\end{subfigure}

\begin{subfigure}{0.49\columnwidth}
\includegraphics[width=1\columnwidth]{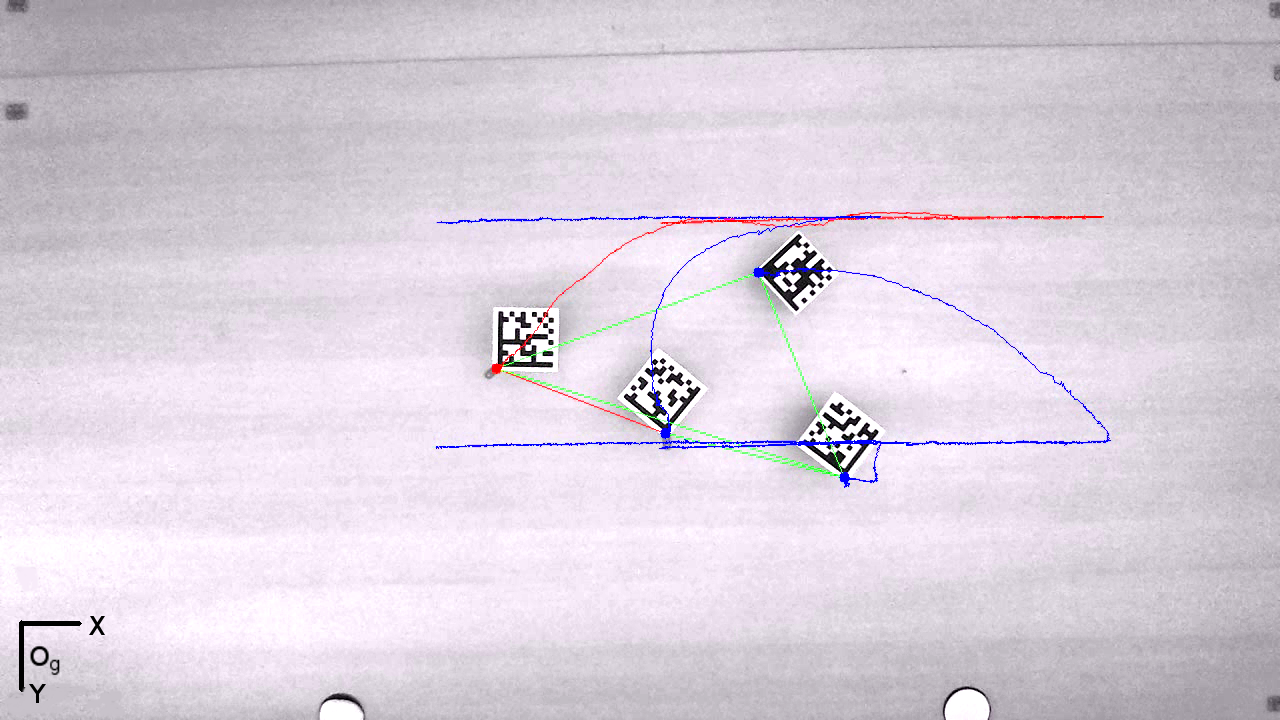}
\caption{Time $t = 195$ secs.}
\end{subfigure}
\begin{subfigure}{0.49\columnwidth}
\includegraphics[width=1\columnwidth]{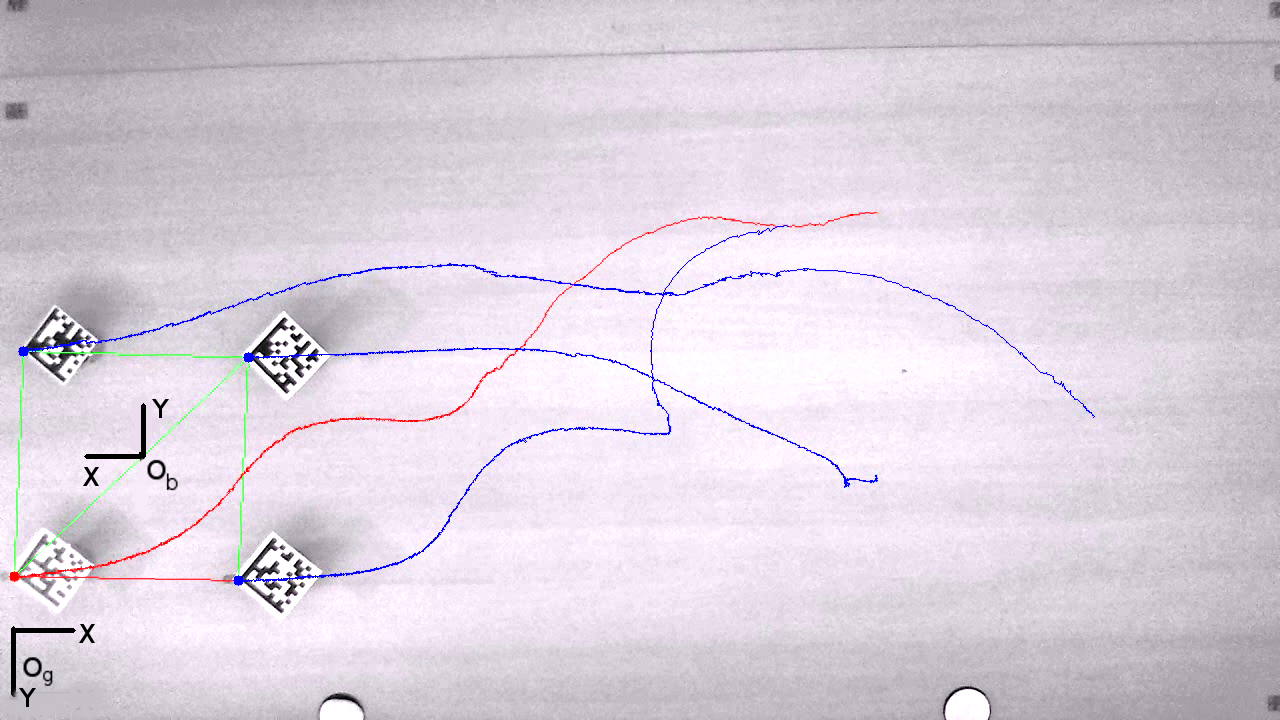}
\caption{Time $t = 283$ secs.}
\end{subfigure}

\begin{subfigure}{0.49\columnwidth}
\includegraphics[width=1\columnwidth]{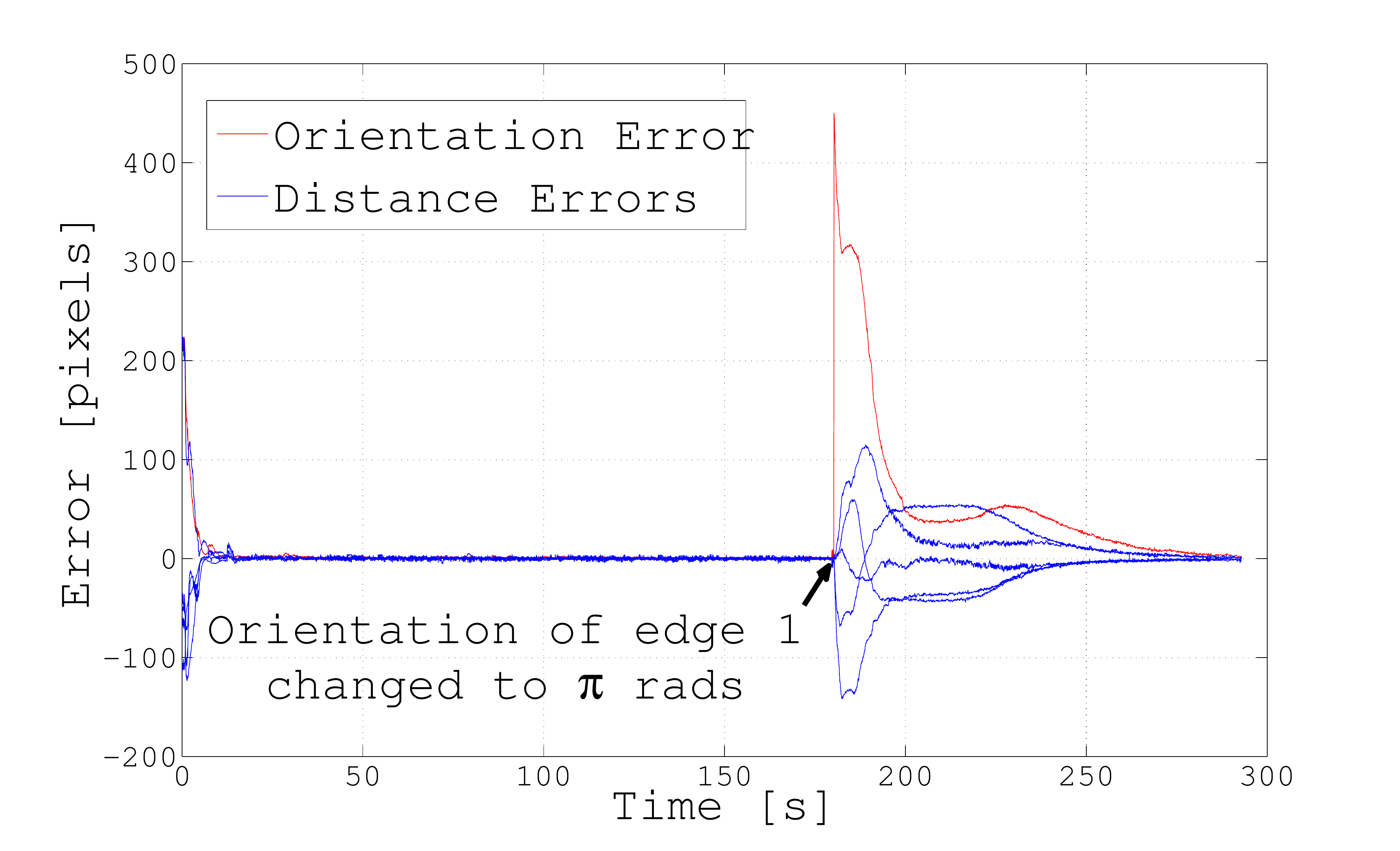}
\caption{}
\end{subfigure}
\begin{subfigure}{0.49\columnwidth}
\includegraphics[width=1\columnwidth]{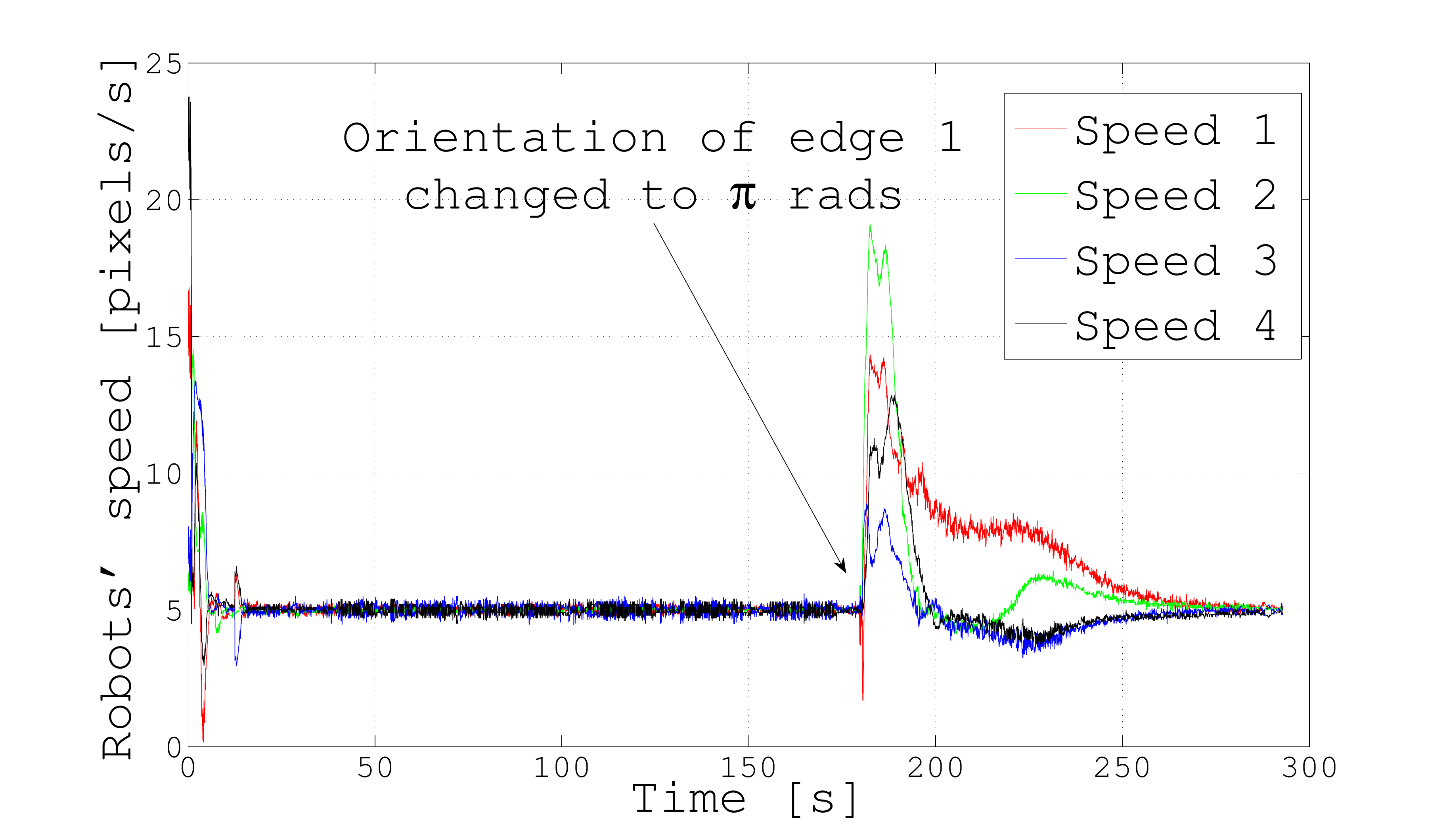}
\caption{}
\end{subfigure}

\caption{Experimental results of a 2D squared formation traveling with 
desired heading and speed using our proposed control law (\ref{eq: controlOr}) with $c=3.5\times10^{-2}$. Note that since we have set $l=1$  in (\ref{eq: Vkquad}), then $D_{\tilde z}$ is not constant and equal to the identity matrix anymore and it should be included as in (\ref{eq: pdynNOmu}). The designed motion parameters have been set in (\ref{eq: misori}). The initial positions of the robots are shown in (a) and the squared configuration of the formation with $0$ rads of heading after $90$ seconds is shown in (b). The error and speed plots corresponding to the experiment are shown in (e) and (f). The desired speed of $5$ pixels/s ($1$ cm/s), is achieved in a short time as it is shown in (f). The orientation leader is marked with a red dot in the video captures, as well as its trajectory. The relative position $^bz_1$ corresponding to the edge $\mathcal{E}_1 = (1,2)$ has been also marked in red, whereas the rest of sensed relative positions have been marked in green. The desired orientation for $z_1^*$ at the beginning is $0$ rads, so $O_g$ and $O_b$ have the same orientation. The steady-state velocity $^b\dot p_c^*$ has been designed to have the same direction as $^bz_1$. Therefore the steady-state translation of the formation is heading to the right in (b). The norm of the orientation error $e_o$ its shown in (e) in red color. At the time $t = 186$ secs, the orientation leader changes the desired orientation of $z_1^*$ to $\pi$ rads, i.e. horizontal motion to the left as it can be noticed in (c) and (d). The video capture (c) shows the rearranging of the agents for the new desired orientation. Finally in (d) at time $t=283$ secs it is shown that the formation achieves the new desired heading of $\pi$ rads along with the desired speed of $5$ pixels/s as it is shown in (f). Note that $O_b$ is only defined once the square is formed.}
\label{fig: he}
\end{figure}

\footnotesize
\begin{equation}
	B =
\begin{bmatrix}
	1 & 0 &-1 & 1 & 0 \\
	-1 & 1 & 0 & 0 & 0 \\
	0 &-1 & 1 & 0 & 1\\
	0 & 0 & 0 &-1 &-1
\end{bmatrix}, \, 
	A =
\begin{bmatrix}
	\mu_1 & 0 &\tilde\mu_3 & \mu_4 & 0 \\
	\tilde\mu_1 & \mu_2 & 0 & 0 & 0 \\
	0 &\tilde\mu_2 & \mu_3 & 0 & \mu_5\\
	0 & 0 & 0 &\tilde\mu_4 &\tilde\mu_5
\end{bmatrix}.
\nonumber
\end{equation}
\normalsize
We designate the robot number $1$ to be the agent that controls the orientation of $z_1$, which is associated to the edge $\mathcal{E}_1 = (1,2)$. We take $l = 1$ for the potential function (\ref{eq: Vkquad}). We implement the control law (\ref{eq: controlOr}) but noting that now $D_{\tilde z}$ is not constant and equal to the identity matrix anymore. The chosen speed for the translation is $5$ pixels/s ($1$cm/s), and we set the desired heading of the translational motion to be the same as the one defined by the direction of $^bz_1$. Because of simplicity and in order to show that Assumption \ref{as: t} is only a sufficient condition, for achieving the mentioned desired steady-state velocity we set the motion parameters as
\begin{align}
\mu = [\begin{smallmatrix}-5 & 0 & 0 & 0 & 5\end{smallmatrix}]^T ,\, \tilde\mu &= [\begin{smallmatrix}-5 & 0 & 0 & 0 & 5\end{smallmatrix}]^T. \label{eq: misori}
\end{align}
At the beginning of the experiment  robot $1$ has for the desired orientation the vector $z_1^* = \begin{bmatrix}225 & 0\end{bmatrix}^T$ pixels. Therefore the desired heading of the formation is $0$ radians with respect to $O_g$. The robots start in random positions close to the origin of $O_g$. During the experiment, once the agent $1$ reaches the imaginary vertical line crossing the position $\begin{bmatrix}1100 & 0\end{bmatrix}^T$ pixels, the value of $z_1^*$ changes to $\begin{bmatrix}-225 & 0\end{bmatrix}^T$ pixels. Therefore the formation will switch to a heading of $\pi$ radians with respect to $O_g$. The gain $c$ in (\ref{eq: controlOr}) has been set to $0.035$. As a remark the chosen $c$ is smaller than the conservative $c^*$ in Theorem \ref{th: ori}. The experimental results are discussed in Figure \ref{fig: he}. 
\begin{remark}
Since all the unicycle robots have the same reference point for the feedback linearization, once the system is at $e=0$ and $e_o = 0$, i.e. all the robots have the same velocity $\dot p_c^*$,  the headings of the robots achieve consensus, i.e. all the robots are pointing at the same direction.
\end{remark}

\subsection{Enclosing and tracking a target experiment}
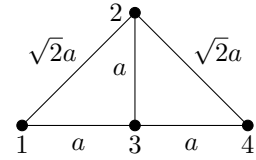
\begin{figure}
	\centering
\begin{tikzpicture}[line join=round]
\filldraw(0,0) circle (2pt);
\filldraw(1.5,1.5) circle (2pt);
\filldraw(1.5,0) circle (2pt);
\filldraw(3,0) circle (2pt);
\draw(0,0)--(1.5,1.5);
\draw(1.5,1.5)--(1.5,0);
\draw(1.5,0)--(0,0);
\draw(3,0)--(1.5,1.5);
\draw(3,0)--(1.5,0);
\node at (0,-.25) {$1$};\node at (.4,1) {$\sqrt{2}a$};\node at (.75,-.25) {$a$};\node at (1.25,1.5) {$2$};\node at (2.25,-.25) {$a$};\node at (2.6,1) {$\sqrt{2}a$};\node at (1.5,-.25) {$3$};\node at (1.3,.75) {$a$};\node at (3,-.25) {$4$};\end{tikzpicture}
	\caption{Desired shape for the three pursuers together with the target. The
	target corresponds to the node number $1$ and $a$ is a scale factor.}
	\label{fig: encs}
\end{figure}

We consider the scenario of having three pursuers and one independent target, set as agent $1$, with constant velocity $\hat v_1 = \begin{bmatrix}-3 & 0.35\end{bmatrix}^T$ pixels/s. The velocity $\hat v_1$ is unknown for the three pursuers. We ask the pursuers to form the shape in Figure \ref{fig: encs} together with the target.
The incidence matrix $B$ corresponding to the sensing topology for the shape in 
Figure \ref{fig: encs} and the derived matrix $A(\mu,\tilde\mu)$ are

\begin{figure}
\centering
\begin{subfigure}{0.49\columnwidth}
\includegraphics[width=1\columnwidth]{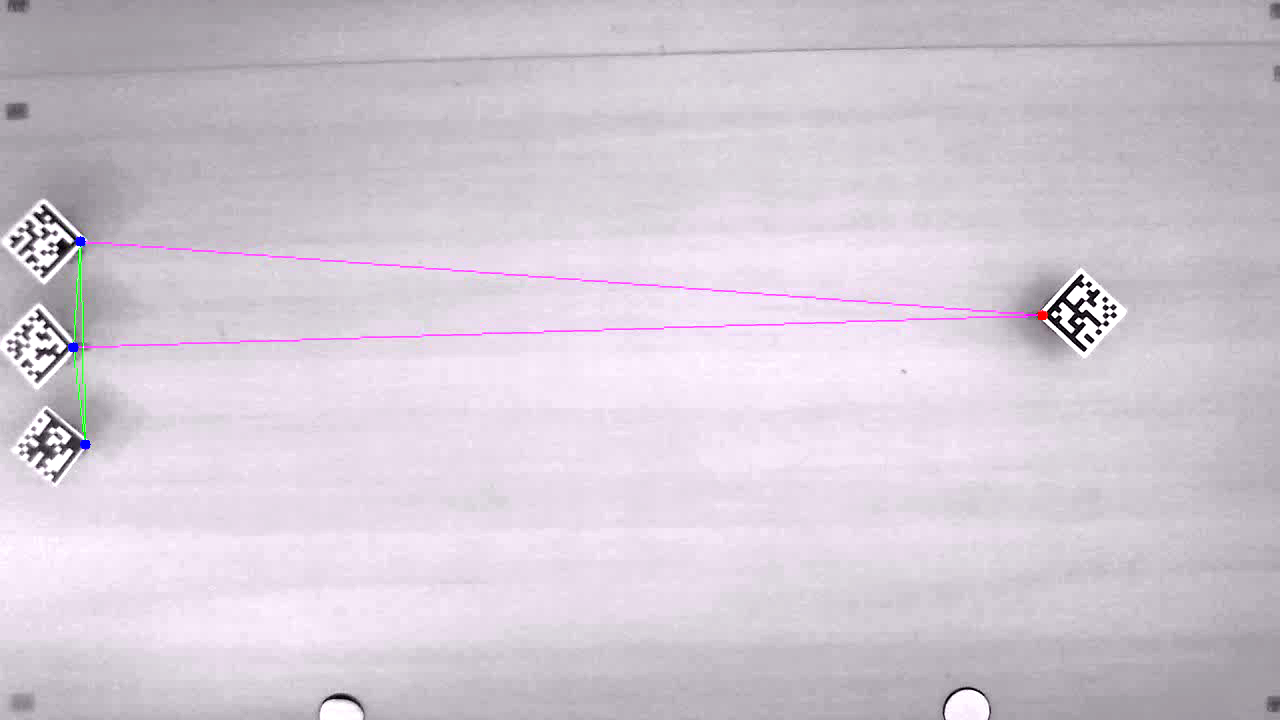}
\caption{Time $t = 0$ secs.}
\end{subfigure}
\begin{subfigure}{0.49\columnwidth}
\includegraphics[width=1\columnwidth]{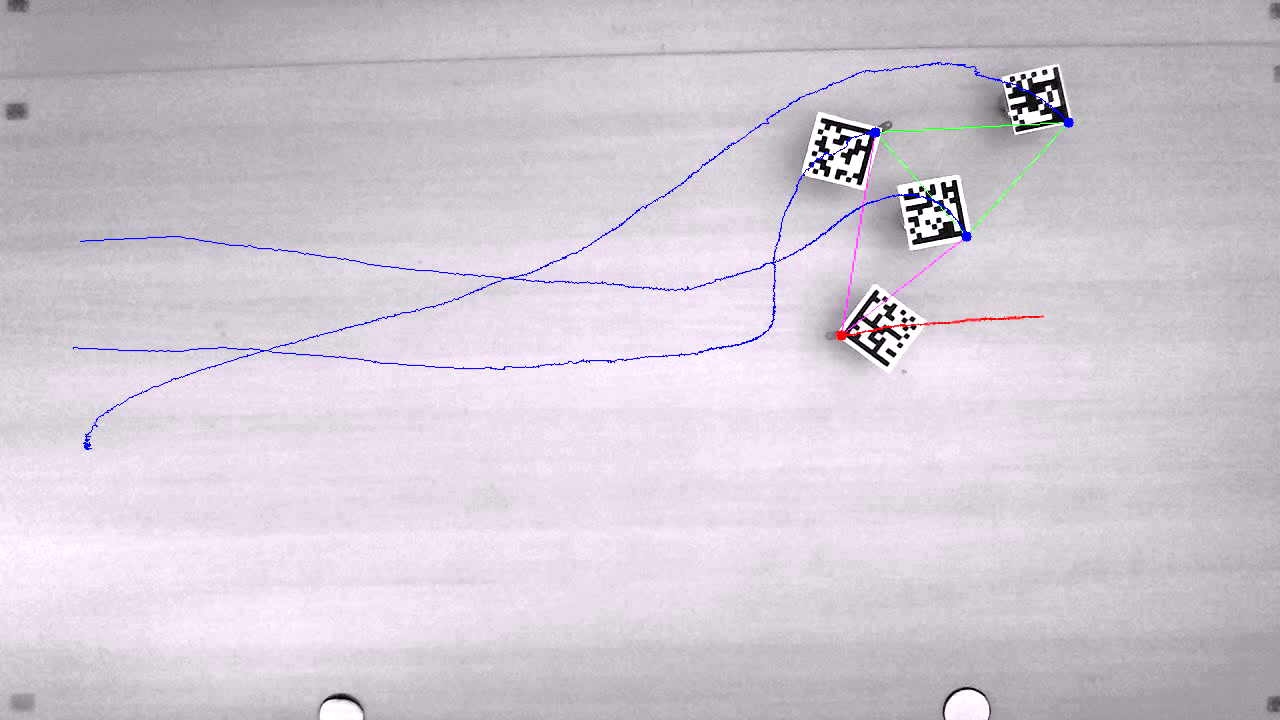}
\caption{Time $t = 68$ secs.}
\end{subfigure}

\begin{subfigure}{0.49\columnwidth}
\includegraphics[width=1\columnwidth]{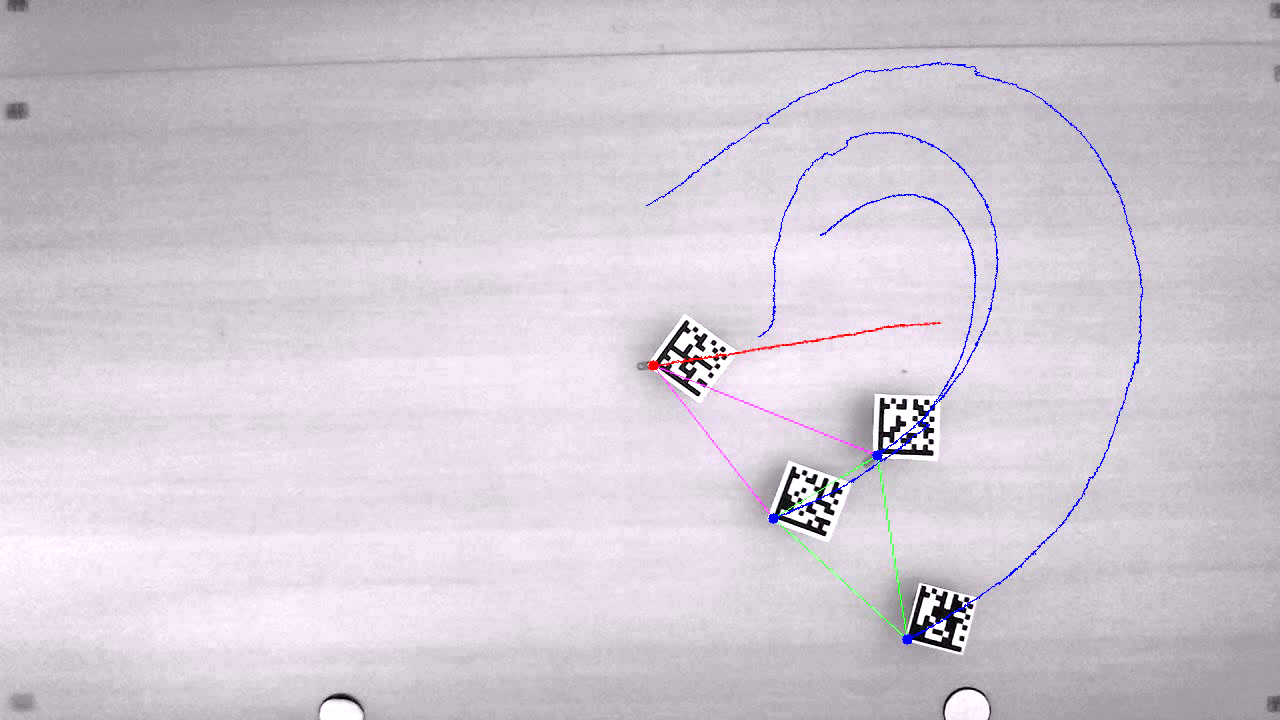}
\caption{Time $t = 135$ secs.}
\end{subfigure}
\begin{subfigure}{0.49\columnwidth}
\includegraphics[width=1\columnwidth]{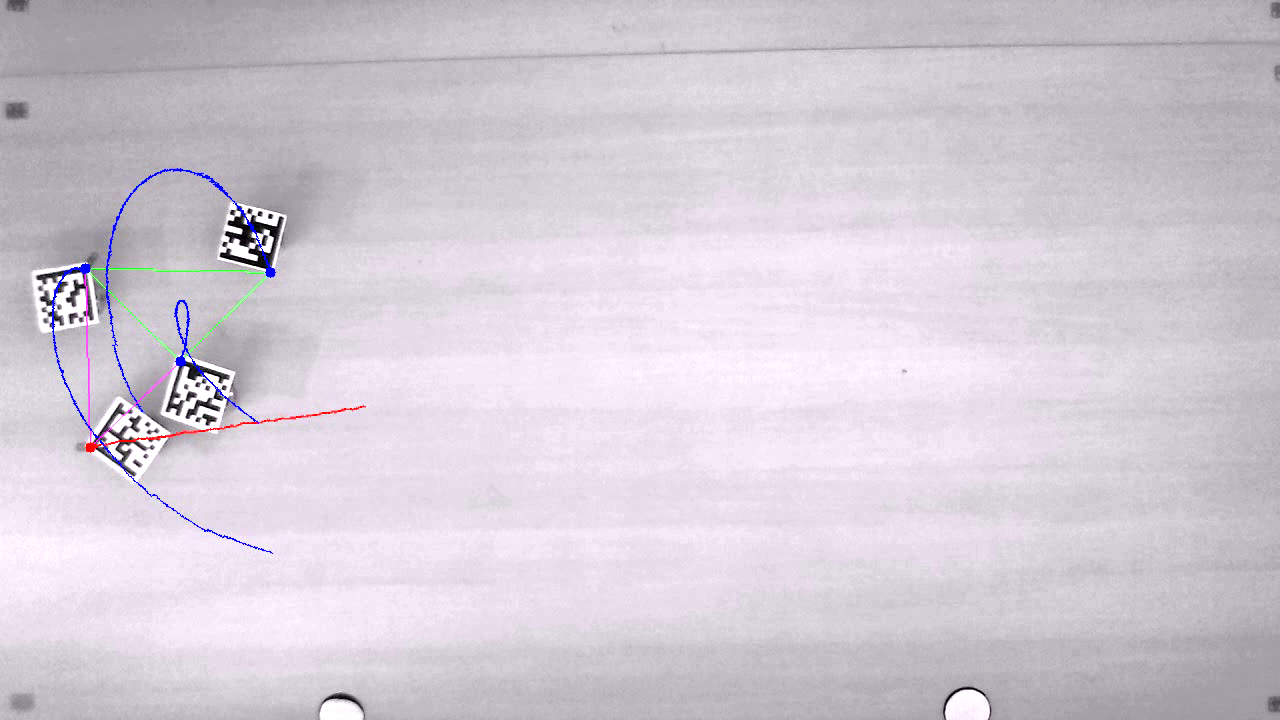}
\caption{Time $t = 335$ secs.}
\end{subfigure}

\begin{subfigure}{0.49\columnwidth}
\includegraphics[width=1\columnwidth]{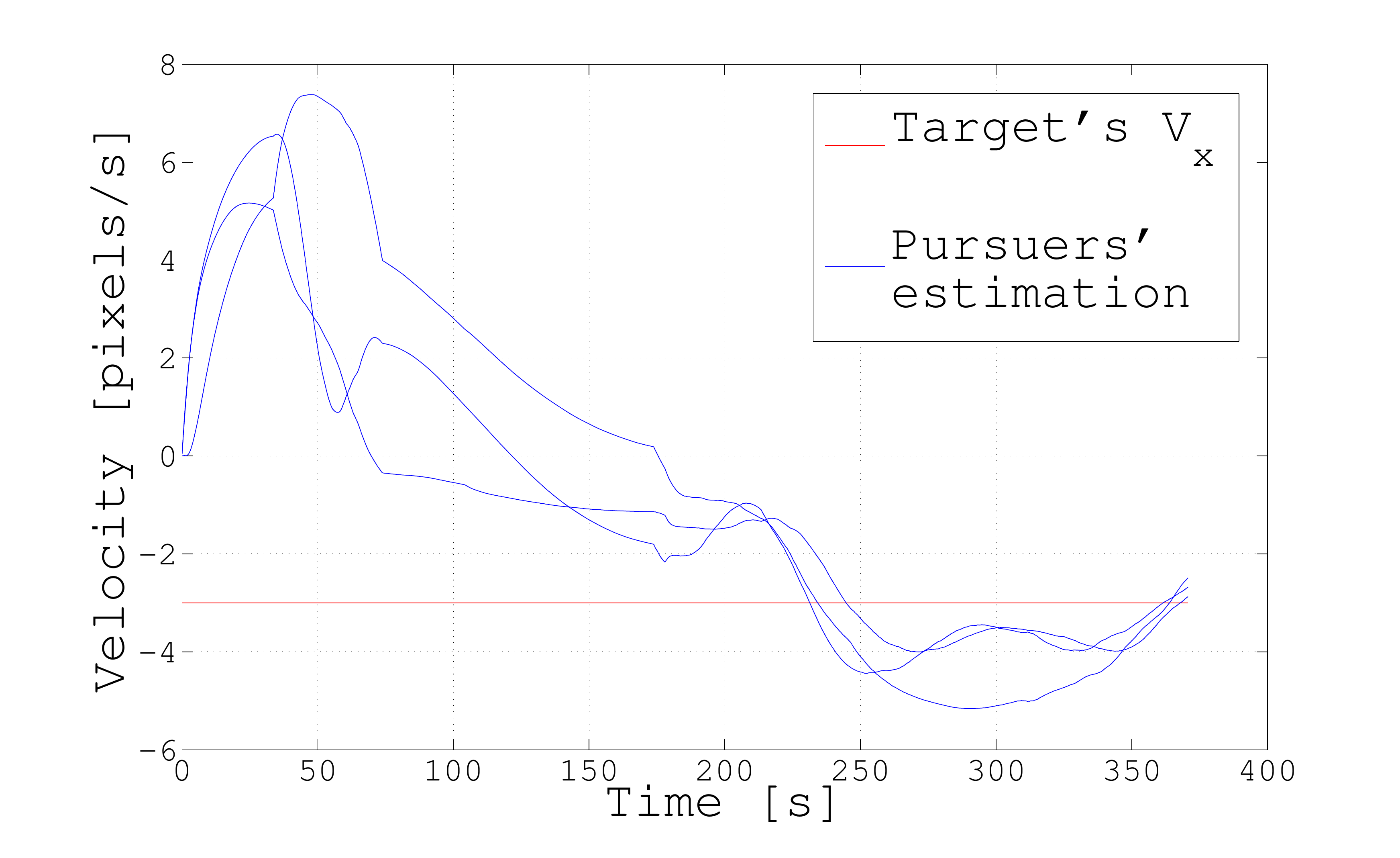}
\caption{}
\end{subfigure}
\begin{subfigure}{0.49\columnwidth}
\includegraphics[width=1\columnwidth]{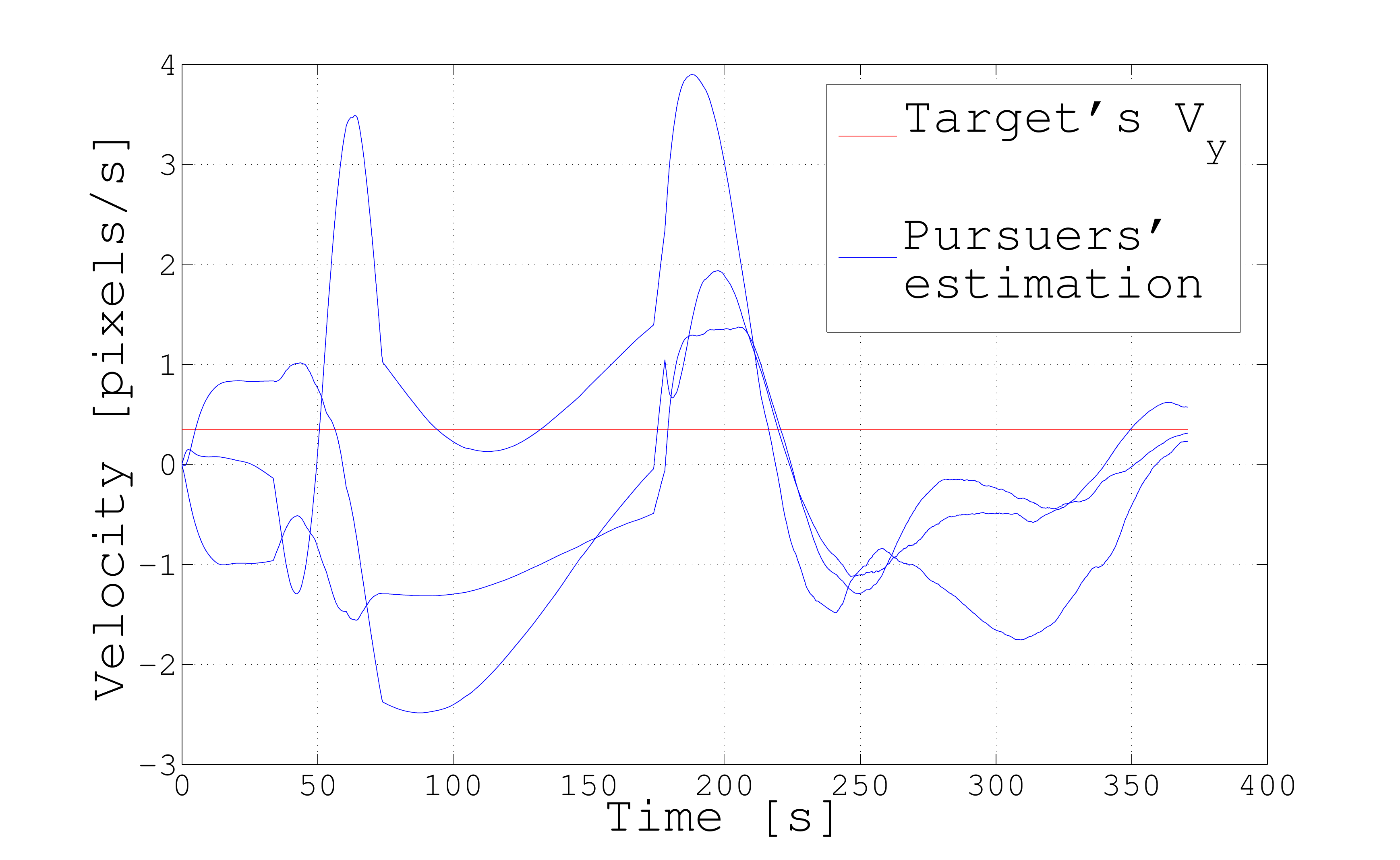}
\caption{}
\end{subfigure}

\begin{subfigure}{0.49\columnwidth}
\includegraphics[width=1\columnwidth]{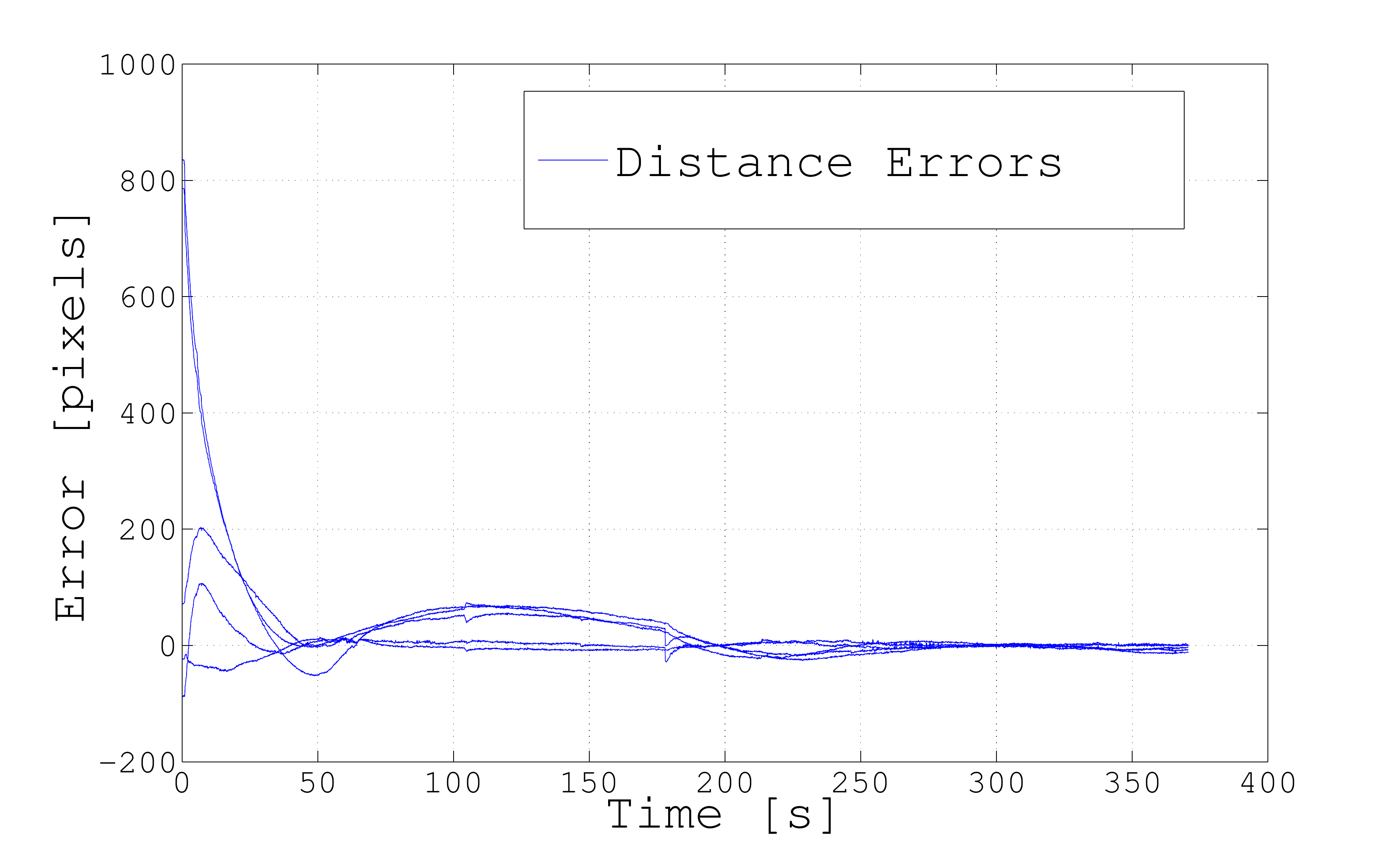}
\caption{}
\end{subfigure}
\begin{subfigure}{0.49\columnwidth}
\includegraphics[width=1\columnwidth]{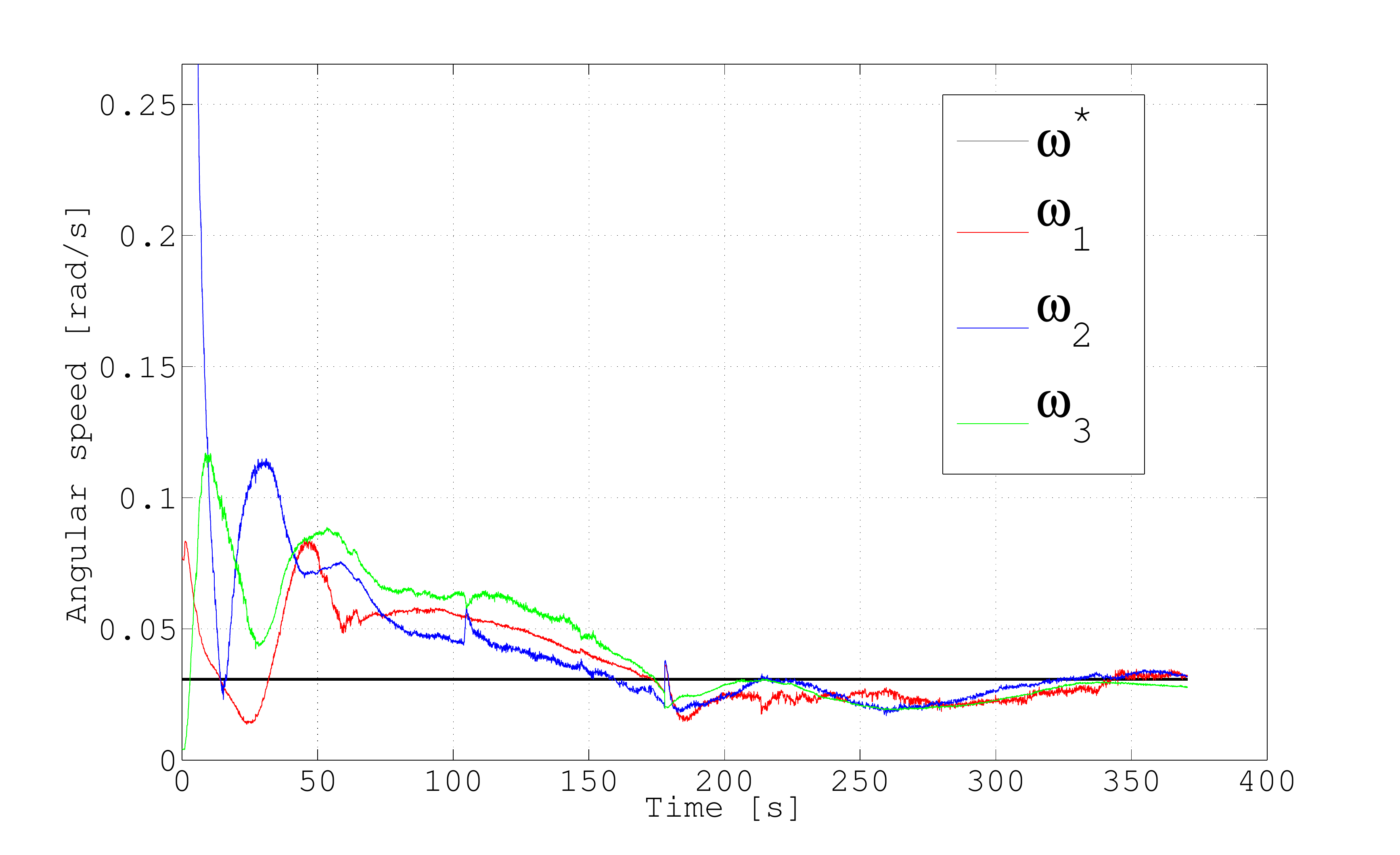}
\caption{}
\end{subfigure}

\caption{Experimental results of a team of three pursuers tracking and enclosing an independent target. The pursuers start at the left of the scene whereas the target starts at the right in (a). The pursuers are tagged with blue dots and the target with a red dot as well as their trajectories. The sensed relative positions are marked with colors, the pink ones concern the target and the green ones concern only the pursuers. Note that there is a pursuer that does not track the target at all. The pursuers have two tasks. The first one is to form the prescribed shape with the target as the one shown in Figure \ref{fig: encs} with the side lengths defined in (\ref{eq: encd}). The second one is to orbit around the target with an angular speed $||\omega^*|| = 0.0308$ rads/s. The designed motion parameters employed for satisfying such requirements are in (\ref{eq: misenc}). In addition the target travels with a constant velocity $\hat v_1 = [-3 \quad 0.35]^T$ pixels/sec, which is unknown to the pursuers. We implement to the pursuers the control laws (\ref{eq: esti}) and (\ref{eq: penc}) with $l = 1$ for the potential (\ref{eq: Vkquad}), therefore $D_{\tilde z}$ is not constant anymore. We have set the control gains $c$ and $\kappa$ to $1\times 10^{-1}$ and  $1\times 10^{-2}$ respectively in (\ref{eq: penc}) and (\ref{eq: esti}). The gain $\kappa$ has been set low in order to prevent the E-pucks to do not reach the saturation level in their velocities. We show the estimation of the two components of $\hat v_1$ by the pursuers in (e) and (f). The asymptotic convergence of the estimated $v$ is slow due to the low values of $\kappa$ and $c$. The convergence of the distance errors to zero in (g) with the estimation of $\hat v_1$, lead for the convergence of the angular speed of the pursuers around the target to $||\omega^*||$ as it is shown in (h). It is shown from (b) to (d) how the pursuers enclose the target and circumnavigate it with the prescribed shape shown in Figure \ref{fig: encs}.}
\label{fig: enc}
\end{figure}

\footnotesize
\begin{equation}
	B =
\begin{bmatrix}
	1 & 0 &-1 & 0 & 0 \\
	-1 & 1 & 0 & -1 & 0 \\
	0 &-1 & 1 & 0 & -1\\
	0 & 0 & 0 & 1 & 1
\end{bmatrix}, \,
	A =
\begin{bmatrix}
	\tilde\mu_1 & 0 & \tilde\mu_3 & 0 & 0 \\
	\mu_1 & \mu_2 & 0 & \tilde\mu_4 & 0 \\
	0 &\tilde\mu_2 & \mu_3 & 0 & \tilde\mu_5\\
	0 & 0 & 0 & \mu_4 & \mu_5
\end{bmatrix}
. \nonumber
\end{equation}
\normalsize
We ask  the pursuers to maintain the following set of inter-agent distances
\begin{equation}
d = \begin{bmatrix}\sqrt{2}a & a & a & \sqrt{2}a & a\end{bmatrix}^T,
	\label{eq: encd}
\end{equation}
where $a = 130$ pixels. We choose $l=1$ for the potential (\ref{eq: Vkquad}) and  implement the controls (\ref{eq: esti}) and (\ref{eq: penc}) to the robots. Note that again since $l=1$, we do not have $D_{\tilde z}$ being constant and equal to the identity anymore. The motion parameters for orbiting around the target $p_1$ are
\begin{equation}
\mu = [\begin{smallmatrix}0 & 0 & 0 & -2a\gamma\sqrt{2} & 2a\gamma\end{smallmatrix}]^T, \,
\tilde\mu = [\begin{smallmatrix}0 & a\gamma & 0 & -a\gamma\sqrt{2} & 0\end{smallmatrix}]^T, \label{eq: misenc}
\end{equation}
where $\gamma\in\mathbb{R}$. Note that $\mu_1$ and $\tilde\mu_3$ are zero since they correspond to the target. The agents $2$ and $3$ are only using $\tilde\mu_4$ and $\tilde\mu_2$ respectively since $z_4^*$ and $z_2^*$ are perpendicular to $z_1^*$. We set the desired steady-state angular speed to $0.038$ rads/s, therefore from (\ref{eq: ome}) the value of $\gamma$ is
\begin{equation}
	||\omega^*|| = \frac{||^b\dot p^*_3||}{||^bp^*_1 - {^bp^*_3}||} = \frac{a\gamma}{a} = \gamma = 0.038 \text{rads/s}. \nonumber
\end{equation}
We have set the control gains $\kappa$ and $c$ to $0.01$ and  $0.1$ respectively in (\ref{eq: esti}) and (\ref{eq: penc}). The gain $\kappa$ has been chosen small in order to prevent saturation in the robots. The chosen $c$ is smaller than $c^*$ in $\mathcal{Q}$, showing the conservative result of $c^*$ in Theorem \ref{th: enc}. The initial values $\hat v_i(0)$ for $i\in\{2, 3, 4\}$ of the estimators have been set to zero. The experimental results are discussed in Figure \ref{fig: enc}. Note that despiting of the local result in Theorem \ref{th: enc}, the initial conditions $p(0)$ and $\hat v(0)$ have been set relatively far away from the desired equilibrium.

\section{Conclusions}
\label{sec: con}
In this paper we have presented a distributed motion controller for rigid formations, which is compatible with the popular gradient-based controllers. The proposed controller allows us to design the steady-state formation motion in the desired shape with respect to a frame attached to the rigid formation. The steady-state velocity vector can be decomposed into two independent terms: one for the constant translation of the centroid of the formation and the other for a constant rotational formation motion around its centroid. Such design has been achieved by motion parameters in the prescribed distances between the agents. Two main applications based on the motion controller have been presented: the first is controlling the heading of a pure translational formation motion and  the second is addressing the problem of enclosing and tracking a target. Experiments with mobile robots in $\R^2$ have been presented in order to validate the proposed applications. We are currently working on extending the mathematical system model to accommodate  agent dynamics modeled by double-integrators. We are also interested in testing our presented results in 3D scenarios involving flying drones.

\ifCLASSOPTIONcaptionsoff
  \newpage
\fi


\bibliographystyle{IEEEtran}
\bibliography{hector_ref}

%




\begin{IEEEbiography}[{\includegraphics[width=1in,height=1.25in,clip,keepaspectratio]{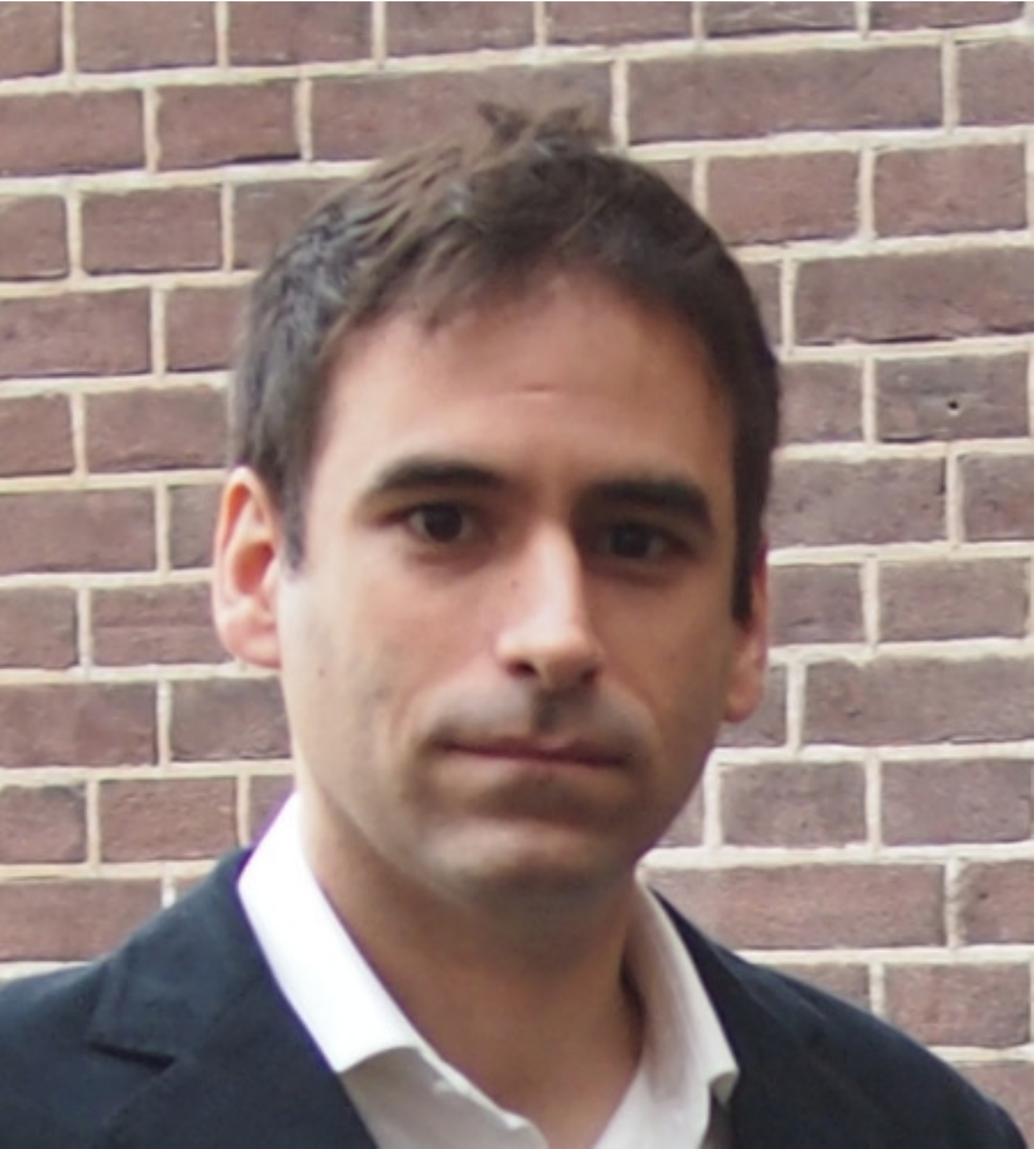}}]{Hector G. de Marina}
	received the M.Sc. degree in electronics engineering from Complutense University of Madrid, Madrid, Spain in 2008, the M.Sc. degree in control engineering from the University of Alcala de Henares, Alcala de Henares, Spain in 2011, and the Ph.D. degree in applied mathematics from the University of Groningen, the Netherlands. He is currently a postdoctoral researcher in the drones group at the School of Civil Aviation (ENAC), Toulouse, France. His research interests include formation control and navigation for autonomous robots.
\end{IEEEbiography}
\begin{IEEEbiography}[{\includegraphics[width=1in,height=1.25in,clip,keepaspectratio]{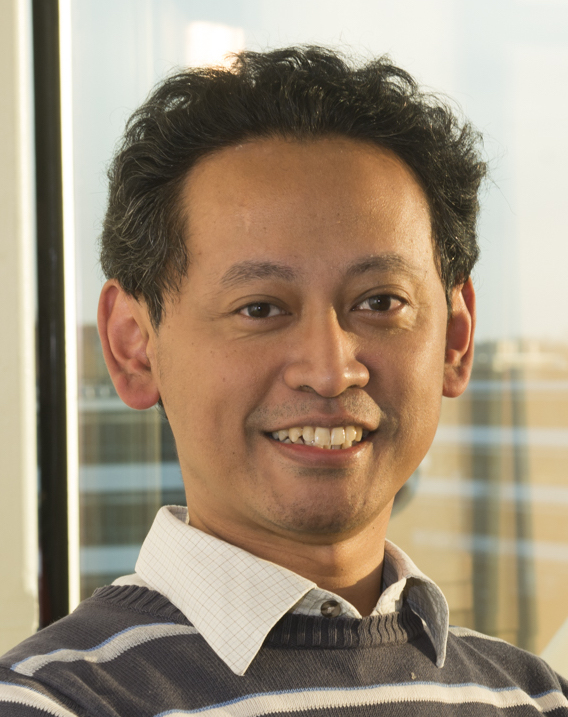}}]{Bayu Jayawardhana}
(SM’13) received the B.Sc. degree in electrical and electronics engineering from the Institut Teknologi Bandung, Bandung, Indonesia, in 2000, the M.Eng. degree in electrical and electronics engineering from the Nanyang Technological University, Singapore, in 2003, and the Ph.D. degree in electrical and electronics engineering from Imperial College London, London, U.K., in 2006. Currently, he is an associate professor in the Faculty of Mathematics and Natural Sciences, University of Groningen, Groningen, The Netherlands. He was with Bath University, Bath, U.K., and with Manchester Interdisciplinary Biocentre, University of Manchester, Manchester, U.K. His research interests are on the analysis of nonlinear systems, systems with hysteresis, mechatronics, systems and synthetic biology. 
He is a Subject Editor of the International Journal of Robust and Nonlinear Control, an associate editor of the European Journal of Control, and a member of the Conference Editorial Board of the IEEE Control Systems Society.
\end{IEEEbiography}
\begin{IEEEbiography}[{\includegraphics[width=1in,height=1.25in,clip,keepaspectratio]{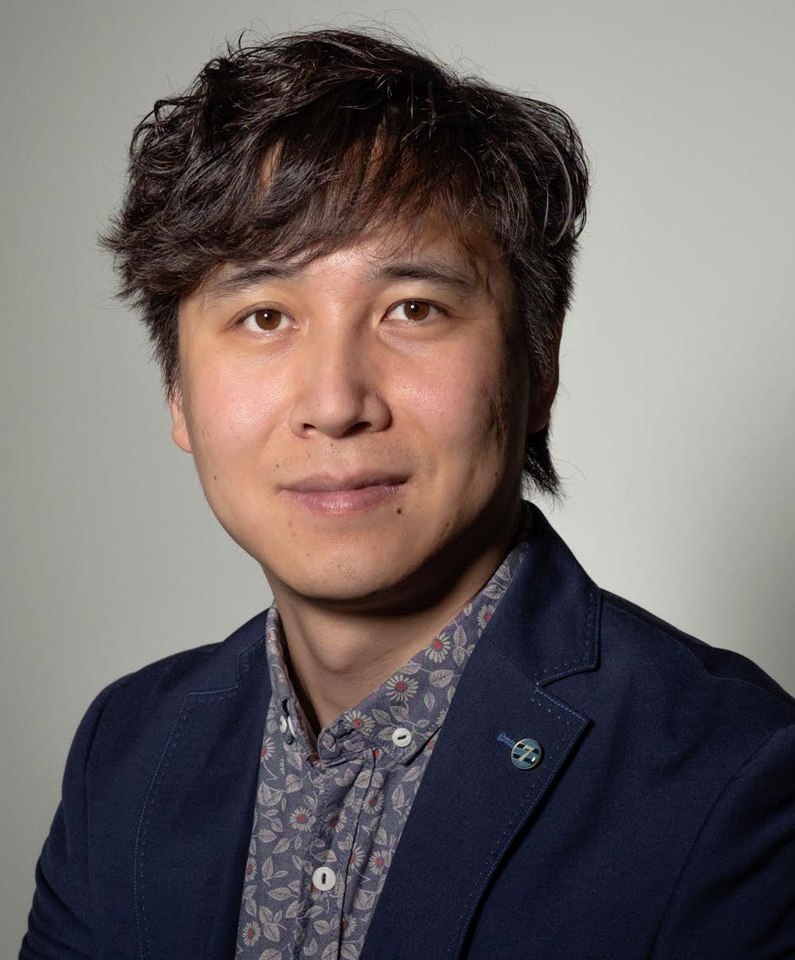}}]{Ming Cao}
	received the PhD degree in electrical engineering from Yale University, New Haven, CT, USA in 2007. He is an associate professor with tenure responsible for the research direction of network analysis and control at the University of Groningen, the Netherlands. His main research interest is in autonomous agents and multi-agent systems, mobile sensor networks and complex networks.  He is an associate editor for \emph{IEEE Transactions on Circuits and Systems II}, \emph{Systems and Control Letters}, and for the Conference Editorial Board of the IEEE Control Systems Society. He is also a member of the IFAC Technical Committee on Networked Systems.
\end{IEEEbiography}

\end{document}